\documentclass[10pt,twocolumn,letterpaper]{article}

\usepackage[pagenumbers]{wacv} 

\definecolor{wacvblue}{rgb}{0.21,0.49,0.74}
\usepackage[pagebackref,breaklinks,colorlinks,allcolors=wacvblue]{hyperref}

\def\wacvPaperID{3178} 
\def\confName{WACV}
\def\confYear{2026}

\newcommand{\method}{\textnormal{DCText}}
\newcommand{\boldmethod}{\textrm{\textbf{DCText}}}

\usepackage{amsmath}
\usepackage{kotex}
\usepackage{amssymb}
\usepackage{booktabs, multirow} 
\usepackage{soul}
\usepackage{xcolor,colortbl} 
\usepackage{changepage,threeparttable} 
\usepackage{tabularx}  
\usepackage{adjustbox} 
\usepackage{enumitem}
\usepackage{bm}
\usepackage{algorithm}
\usepackage{algpseudocode}

\usepackage[accsupp]{axessibility}  

\title{DCText: Scheduled Attention Masking for Visual Text Generation via Divide-and-Conquer Strategy}

\author{Jaewoo Song$^{1,2}$~~~~~~~~~~Jooyoung Choi$^{1}$~~~~~~~~~~Kanghyun Baek$^{3}$~~~~~~~~~~Sangyub Lee$^{3}$\\
Daemin Park$^{1}$~~~~~~~~~~Sungroh Yoon$^{1,3,4,}$\footnotemark[1]\\
$^1$Department of Electrical and Computer Engineering, Seoul National University\\
$^2$Global Technology Research, Samsung Electronics\\
$^3$IPAI, $^4$AIIS, ASRI, INMC, ISRC, Seoul National University\\
{\tt\small \{woo.song, jy\_choi, qor6271, nickyub, eoalsqkr12, sryoon\}@snu.ac.kr}
}

\begin{document}

\twocolumn[{
\renewcommand\twocolumn[1][]{#1}
\maketitle
\begin{center} 
\centering
    \includegraphics[width=0.95\textwidth]{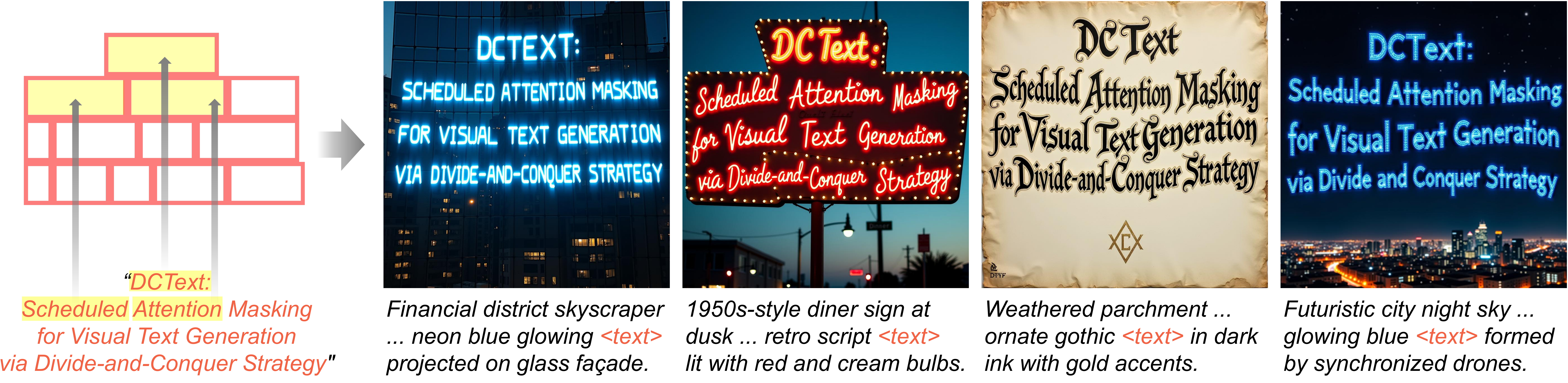}
    \captionof{figure}{Given a global prompt and target regions (red boxes), \method{} decomposes the target text (highlighted in red) and assigns it to regions, enabling accurate and coherent visual text generation, which the base Flux~\cite{flux1-dev} model struggles to handle reliably. The prompts below each image are abbreviated from the original global prompts (full prompts in Appendix~\ref{sec-e3:Abbreviated Prompts}).}
    \label{fig:thumbnail}
\end{center}
}]

\renewcommand{\thefootnote}{\fnsymbol{footnote}}
\footnotetext[1]{Correspondence to: Sungroh Yoon (sryoon@snu.ac.kr)}

\begin{abstract}

Despite recent text-to-image models achieving high-fidelity text rendering, they still struggle with long or multiple texts due to diluted global attention. We propose DCText, a training-free visual text generation method that adopts a divide-and-conquer strategy, leveraging the reliable short-text generation of Multi-Modal Diffusion Transformers.
Our method first decomposes a prompt by extracting and dividing the target text, then assigns each to a designated region.
To accurately render each segment within their regions while preserving overall image coherence, we introduce two attention masks—Text-Focus and Context-Expansion—applied sequentially during denoising.
Additionally, Localized Noise Initialization further improves text accuracy and region alignment without increasing computational cost.
Extensive experiments on single- and multi-sentence benchmarks show that DCText achieves the best text accuracy without compromising image quality while also delivering the lowest generation latency.
\end{abstract}

\section{Introduction}
\label{sec:intro}

Visual text generation has consistently been a challenging task within the text-to-image (T2I)~\cite{ldm,sd3,sdxl,imagen,flux1-dev} domain. Despite the impressive image quality of early T2I models, they still struggle to generate accurate and natural-looking visual text.
While many methods~\cite{liu2022character, tuo2023anytext, yang2023glyphcontrol, chen2023textdiffuser, liu2024glyph}
attempt to overcome this limitation by adding auxiliary modules or fine-tuning, they are often constrained by the capacity of the base model, resulting in poor quality.

Recently, Multi-Modal Diffusion Transformer (MM-DiT) models such as Stable Diffusion 3 series~\cite{sd3} and Flux~\cite{flux1-dev}, equipped with powerful text encoders~\cite{raffel2020exploring}, have significantly improved text rendering capabilities inherently. Building on this, several approaches~\cite{hu2025amosamplerenhancingtext, du2025textcrafter} directly leverage them as backbones in a training-free manner to improve visual text generation.
However, these methods still introduce inference-time computational overhead and lack sufficient text accuracy.
In particular, the latter issue primarily stems from the global attention mechanism: as the amount of text to be rendered increases, the full attention structure in MM-DiT—where all text and image tokens mutually attend to one another—dilutes attention to individual text tokens,
thereby causing omissions, typos, and misplacements that degrade the overall text fidelity.

In this paper, we introduce \method{}, which addresses the limitations of the full attention by applying scheduled attention masking to ensure accurate text generation at specified positions. Our method builds on the observation that Flux can reliably produce relatively short, single pieces of text with high fidelity. Leveraging this, we adopt a divide-and-conquer strategy: 
a global prompt containing multiple or lengthy rendering texts is decomposed into partial prompts based on its text content. 
Each partial prompt is then responsible for generating its assigned text within a designated region, while the overall image remains consistent with the global prompt.
This divide-and-conquer generation is implicitly carried out within a single denoising process, enabled solely by attention masking.
However, this approach requires careful mask design, as the split text segments belong to both partial and global prompts, increasing the risk of duplicated text generations, and the image regions generated by these separated prompts often struggle to blend seamlessly with the surrounding background.

To address these challenges, we design two attention masks. \emph{Text-Focus Attention Mask}: concentrates attention into each designated region, ensuring that the generated text is accurate and confined to the target region without duplication. \emph{Context-Expansion Attention Mask}: allows bidirectional interactions between each region and its surrounding background, enabling smooth and coherent transitions across boundaries.
By sequentially applying these masks during the denoising process, our method balances accurate text fidelity and visual coherence in the final output.
In addition, we propose a \emph{Localized Noise Initialization} approach, which refines the initial noise to provides spatial guidance for the text to be rendered in each region. This approach is not only computationally efficient but also improves region–text alignment and enhances the text accuracy.

We evaluate \method{} on diverse datasets containing both single and multiple rendering sentences. Our method achieves the highest text accuracy among other tuning-free approaches, while also delivering the best image quality. Remarkably, these results are obtained with the fewest denoising steps and the lowest latency, and \method{} further provides flexible controllability over text placement.
\section{Related Work}

\subsection{Visual Text Generation}
With advances in text-to-image models, various methods have been proposed to generate images containing text.
Some approaches train character-level text encoders~\cite{liu2022character, zhao2024udifftext, wang2025dreamtext, liu2024glyph, liu2024glyph2, wang2025designdiffusion}, enabling models to recognize and generate glyph structures at the character level.
Others incorporate external modules into T2I models~\cite{yang2023glyphcontrol, tuo2023anytext, wang2024textmastero, ma2023glyphdraw, zeng2024textctrl, wang2024high, chen2023textdiffuser, chen2024textdiffuser} to handle glyph-level information, which not only improves text rendering accuracy but also enables spatial control over text layout through masks or region-based guidance. However, these methods are built upon U-Net–based Stable Diffusion models~\cite{rombach2022high} and often require additional training, exhibiting limitations in both visual fidelity and text accuracy.


To overcome these challenges, recent research has focused on training-free approaches that leverage the pre-trained capabilities of Flux to enhance text generation at inference time.
AMO Sampler~\cite{hu2025amosamplerenhancingtext} improves text rendering accuracy by introducing a stochastic sampler along with a mask based on cross attention, but lacks explicit control over text positioning.
TextCrafter~\cite{du2025textcrafter} addresses the multi-text rendering task by separately denoising layout regions where text is generated and re-weighting attention maps to mitigate text blurring.
But it often struggles to achieve seamless integration between text regions and the background, resulting in unnatural visual transitions.
Furthermore, both methods still suffer from limited text accuracy and inference-time computational overhead.

\subsection{Attention Control}
Attention maps have been widely explored for controlling generation in diffusion models.
Prompt-to-Prompt~\cite{hertz2022prompt} reveals that cross-attention maps reflect spatial alignment between prompt and image, which can be exploited for prompt-based image editing by replacing or re-weighting these maps.
Additional techniques compute loss signals from attention maps and apply latent updates to better align outputs with textual prompts~\cite{chefer2023attend, syngen, a_star}.
Other approaches extend this idea to incorporate layout conditioning, enabling models to reflect both prompt semantics and spatial constraints~\cite{xie2023boxdiff, dahary2024yourself}.
However, these methods rely solely on cross-attention mechanism, which limits their flexibility in capturing fine-grained spatial relationships. In addition, the latent update procedures introduce extra memory and computational overhead during inference.

More recently, MM-DiT-based models have adopted joint attention mechanisms, where text and image tokens are concatenated and jointly attended.
Some studies apply spatial masks directly to joint attention maps to enforce layout constraints without latent updates.
For instance, Regional-Prompting~\cite{chen2024training} enables compositional generation by performing both manipulated and unmanipulated attention evaluations, which doubles number of function evaluations (NFE) per denoising step.
In contrast, DreamRenderer~\cite{zhou2025dreamrenderer} replicates image tokens for each instance during attention, leading to substantial overhead within each step, instead of doubling NFE.
Above all, these approaches mainly focus on generic objects while overlooking text, which is more difficult to generate due to strict requirements on glyph accuracy and character sequencing—especially when rendering large amounts of text.

\begin{figure*}[t]
  \centering
    \includegraphics[width=0.9\textwidth]{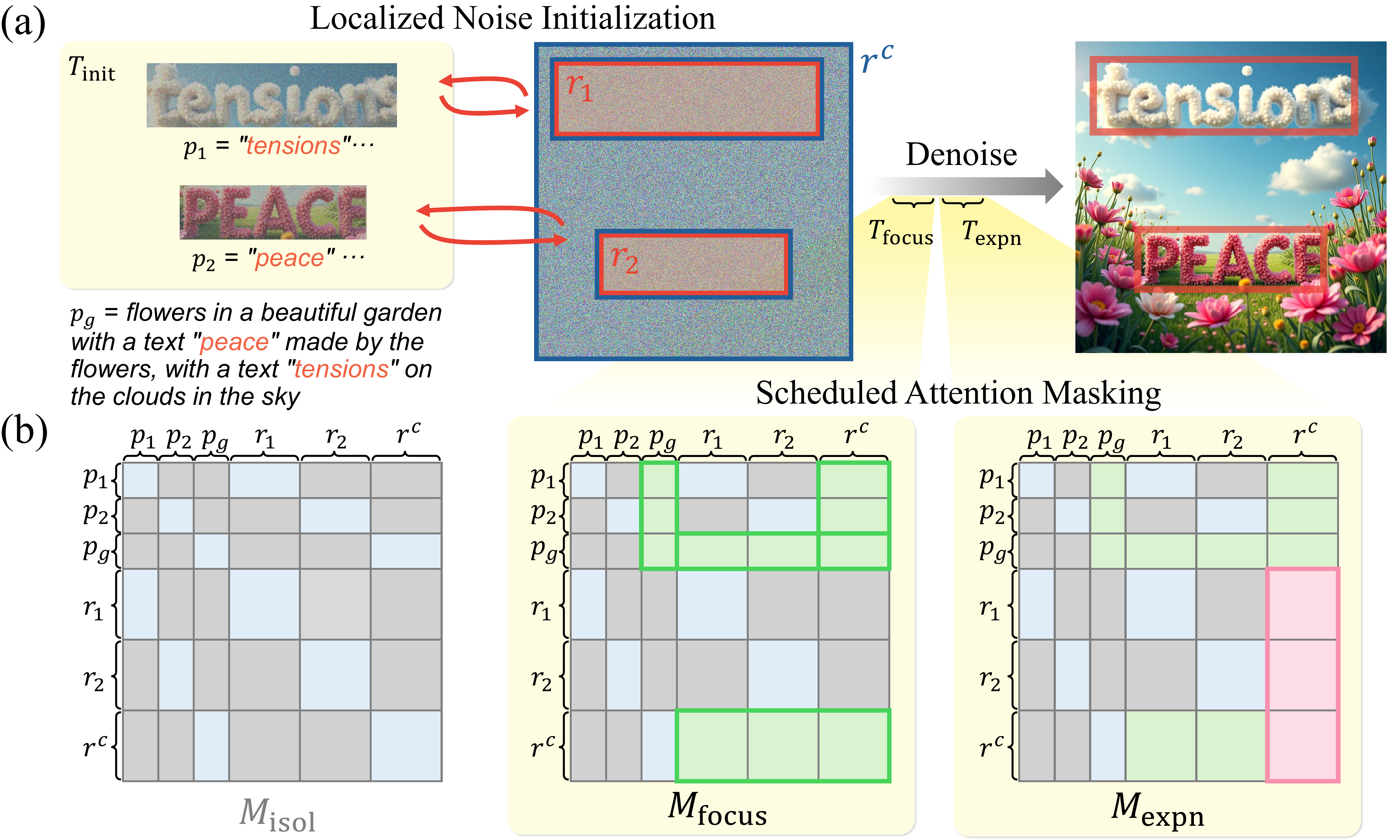}
    \caption{\textbf{Overview of \boldmethod{}.} 
    (a) Textual regions ($r_1, r_2$) are extracted from the initial noise and independently denoised with their textual prompts ($p_1, p_2$) for $T_{\mathrm{init}}$ steps. The refined patches are then blended back to form the localized initial noise. (b) It is then sequentially denoised using two attention masks, $M_{\mathrm{focus}}$ and $M_{\mathrm{expn}}$. Both masks build on $M_{\mathrm{isol}}$, which restricts attention to text and image tokens within each region (blue).
    $M_{\mathrm{focus}}$ introduces four attention areas (green) to allow controlled information flow between textual regions and the background, supporting accurate text rendering.
    $M_{\mathrm{expn}}$ further enables region-to-background attention (pink) to promote smooth transitions. In all three masks, colored areas denote allowed attention (1), while gray areas indicate masked attention (0).}
    \label{fig:mask}
\end{figure*}


\section{Method}
\label{Method}

Our goal is to accurately and naturally generate texts from a global prompt, which may contain long or multiple sentences.
To achieve this, we leverage bounding box conditions: given a set of regions requiring visual text segments (textual regions) and a global prompt,
we extract the target texts embedded in the prompt and divide them to match the number of regions, generating each within its assigned region.
Building on the attention mechanism of the Multi-Modal Diffusion Transformer,
we control the information flow between textual and non-textual content (both across prompts and regions) using carefully designed attention masks.
This approach enables region-centric text generation while ensuring smooth transitions in the background, keeping the overall image coherent with the global prompt.
Our method, \method{}, improves text accuracy by applying these attention masks at inference time, with minimal overhead compared to the base text-to-image model. It requires no training or fine-tuning and avoids heavy computations such as gradient calculations during inference~\cite{a_star, syngen}.

In the following sections, we first review the attention mechanism in MM-DiT (\cref{Preliminary}),
then describe how this attention is regulated by two novel attention masks and their design formulation (\cref{Attention Mask Control}),
next present the strategy for obtaining the initial noise (\cref{Localized Noise Initialization}),
and finally outline the overall \method{} pipeline (\cref{pipeline}).

\subsection{Preliminary: Attention in MM-DiT}
\label{Preliminary}

Recent T2I models~\cite{esser2024scaling, flux1-dev} employ MM-DiT blocks, which operate on a unified token sequence formed by concatenating text and image tokens. Each block computes attention over this combined sequence as follows:
\begin{equation}
\mathrm{Attn}(Q, K, V) = AV, \quad A = \mathrm{softmax}\left( \frac{QK^\top}{\sqrt{d}} \right),
\label{eq:attention}
\end{equation}
where $Q = \mathrm{concat}(Q_{\mathrm{text}}, Q_{\mathrm{img}})$, with $Q_{\mathrm{text}}$ and $Q_{\mathrm{img}}$ representing the queries derived from the text prompt tokens and image patch tokens, respectively; key ($K$) and value ($V$) are constructed in the same manner as $Q$.

During attention computation, it is possible to regulate the information flow between tokens by modifying or masking the attention map $A$. In particular, applying an attention mask $M$ enables selective interaction by suppressing irrelevant connections while preserving meaningful ones:
\begin{equation}
A = \mathrm{softmax}\left( \frac{QK^\top}{\sqrt{d}} + \log(M) \right).
\label{eq:attention_map_with_mask}
\end{equation}
Here, the binary values in $M$ (1 for allowed connections, 0 for masked) are transformed via the $\log$ function into 0 and $-\infty$, respectively, to be added before the softmax operation.

\subsection{Scheduled Attention Masking}
\label{Attention Mask Control}

We first decompose a global prompt, which contains multiple or lengthy target texts to be rendered, into a set of partial prompts (textual prompts) based on the number of textual regions $n$.
For example, given a global prompt such as \textit{`flowers in a beautiful garden with a text ``peace'' made by the flowers, with a text ``tensions'' on the clouds in the sky'} and two regions, 
we construct textual prompts by separating \textit{peace} and \textit{tensions}, and assign each to its region.

These $n$ textual prompts are encoded in the same manner as the global prompt, resulting in $\{p_i\}_{i=1}^n$.
We extend the attention inputs $Q$, $K$, and $V$ in~\cref{eq:attention} to incorporate both the textual prompts and the global prompt as follows:
\begin{equation}
Q_{\mathrm{text}} = \mathrm{concat}(\{ Q_{p_{i}} \}_{i=1}^n, Q_{p_g}),
\label{eq:q_text}
\end{equation}
where $Q_{p_i}$ and $Q_{p_g}$ are queries from the $i$-th textual prompt and the global prompt, respectively.
In parallel, we organize the image-side query $Q_{\mathrm{img}}$ not by extending it, but by decomposing it based on the textual regions.
Specifically, image patches are divided into region-specific parts $\{r_i\}_{i=1}^n$ based on the given bounding boxes, and a background region $r^c = \left( \bigcup_{i=1}^n r_i \right)^c$, \ie the complement of all regions:
\begin{equation}
Q_{\mathrm{img}} = \mathrm{concat}\left( \{ Q_{r_{i}} \}_{i=1}^n,\ Q_{r^c} \right).
\label{eq:q_img}
\end{equation}
$K$ and $V$ are constructed analogously to $Q$.

We thus compute attention over the token sequence consisting of prompts $\{p_i\}_{i=1}^{n+1}$ and regions $\{r_i\}_{i=1}^{n+1}$, where $p_{n+1} = p_g$ and $r_{n+1} = r^c$.
Note that the last prompt $p_g$ serves as a global description that encompasses the textual prompts, while the last region $r^c$ corresponds solely to the background region, excluding all textual regions.

In this setup, we propose two types of attention masks to guide the denoising process for accurate and coherent text generation. 
The text-focus attention mask $M_{\mathrm{focus}}$ biases the overall attention toward textual regions, enabling accurate text rendering at the correct regions while allowing the background to remain contextually natural.
The context-expansion attention mask $M_{\mathrm{expn}}$, on the other hand, further allows each textual region to attend to the background region, promoting smoother transitions and coherence at the boundaries between the region and the background.

To motivate the construction of these two masks, we first introduce the region-isolation attention mask $M_{\mathrm{isol}}$, which enforces strict separation across all regions. While not used in our method, $M_{\mathrm{isol}}$ serves as a conceptual foundation from which our proposed masks are derived. In the following, we detail the formulation of $M_{\mathrm{isol}}$ and then construct $M_{\mathrm{focus}}$ and $M_{\mathrm{expn}}$ by gradually relaxing the strict constraints imposed by $M_{\mathrm{isol}}$. These masks are illustrated in~\cref{fig:mask}b.

\begin{algorithm}[!t]
\caption{\method{} Pipeline}
\label{alg:dctext}

\textbf{Input}: \\
Prompts $\{p_i\}_{i=1}^{n+1}$ (with $p_{n+1}$: global prompt), Regions $\{r_i\}_{i=1}^{n+1}$ (with $r_{n+1}$: background), Denoising steps $T$, $T_{\mathrm{init}}$, $T_{\mathrm{focus}}$, $T_{\mathrm{expn}}$, Blending weight $\alpha$ \\
\textbf{Output}: Final image $x_0$

\begin{algorithmic}[1]
\State Make attention masks $M_{\mathrm{focus}}$, $M_{\mathrm{expn}}$
\State Sample $z_T \sim \mathcal{N}(0, I)$

\Statex
\Comment{Localized Noise Initialization}
\For{$i = 1$ to $n$}
    \State $z^{(i)} \gets$ Extract($z_T$, $r_i$)
    \For{$t$ for $T_{\mathrm{init}}$ steps}
        \State $z^{(i)} \gets$ Denoise($z^{(i)}, p_i$)
    \EndFor
    \State $z_T \gets$ Reinsert($z_T$, $r_i$, $z^{(i)}, \alpha$)
\EndFor
\State $z \gets z_T$

\Statex
\Comment{Text-Focus Denoising}
\For{$t$ for $T_{\mathrm{focus}}$ steps}
    \State $z \gets$ DenoiseWithMask($z$, $\{p_i\}_{i=1}^{n+1}$, $M_{\mathrm{focus}}$)
\EndFor

\Statex
\Comment{Context-Expansion Denoising}
\For{$t$ for $T_{\mathrm{expn}}$ steps}
    \State $z \gets$ DenoiseWithMask($z$, $\{p_i\}_{i=1}^{n+1}$, $M_{\mathrm{expn}}$)
\EndFor

\Statex
\Comment{Global Denoising}
\For{$t$ for $T - (T_{\mathrm{init}} + T_{\mathrm{focus}} + T_{\mathrm{expn}})$ steps}
    \State $z \gets$ Denoise($z$, $p_{n+1}$)
\EndFor

\Statex
\State \Return Decode($z$)
\end{algorithmic}
\end{algorithm}

\begin{figure*}[!t]
  \centering
    \includegraphics[width=0.923\textwidth]{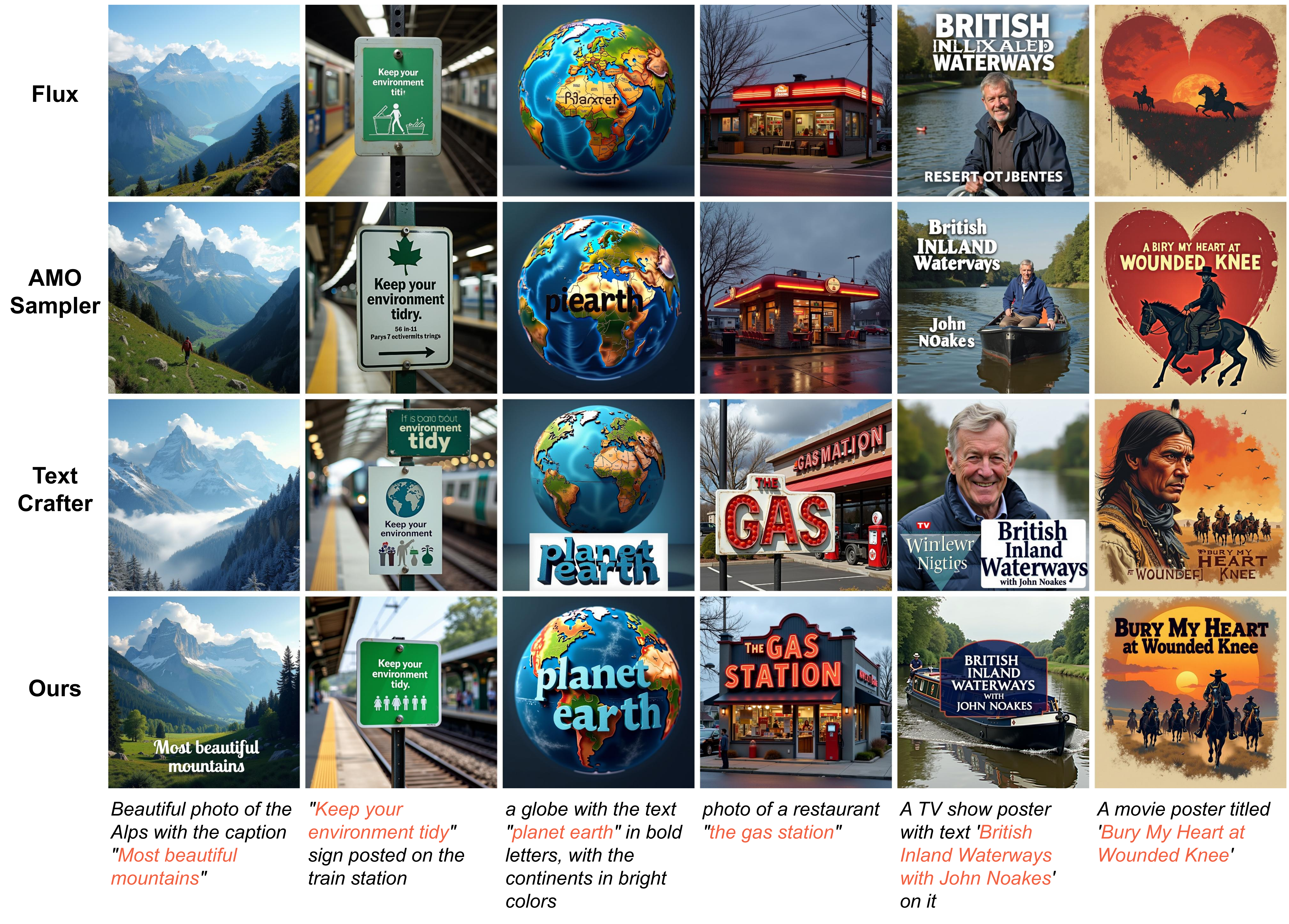} 
    \caption{\textbf{Qualitative comparison on single sentence.} Prompts, including the sentence to be rendered (highlighted in red), are shown below each column. All comparisons are generated from the same initial noise.}
    \label{fig:single}
\end{figure*}

\begin{figure}[!t]
  \centering
    \includegraphics[width=0.95\columnwidth]{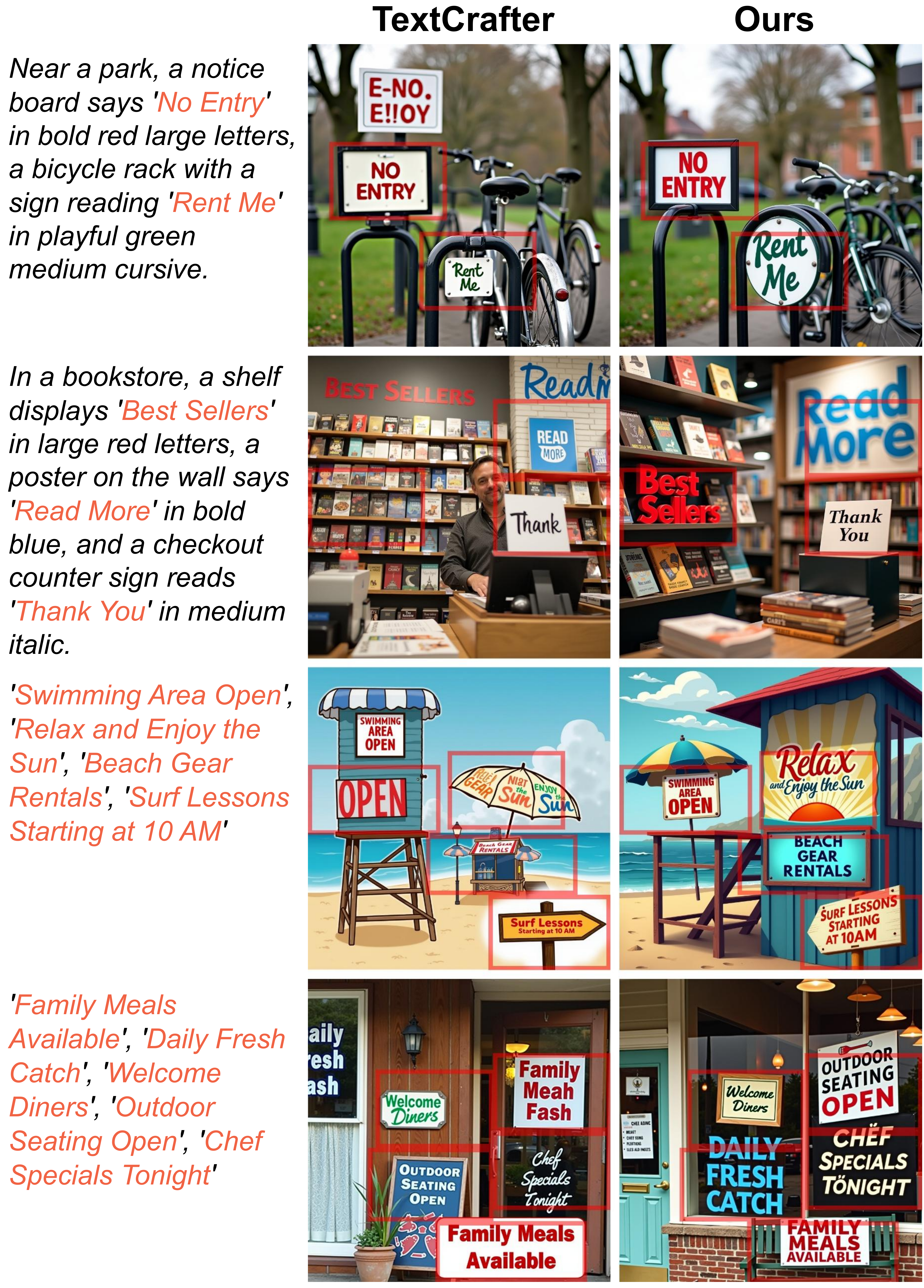}
    \caption{\textbf{Qualitative comparison on multiple sentences.} Comparison of generation results with varying numbers of sentences (2–5) in a single prompt. Sentences and corresponding regions are highlighted in red (only target texts are shown for the two prompts below; full prompts are in Appendix~\ref{sec-e3:Abbreviated Prompts}). Our method consistently renders accurate text in the correct regions.}

    \label{fig:multi}
\end{figure}

\paragraph{Region-Isolation Attention Mask.}
Our divide-and-conquer approach begins by partitioning the task, enabling each textual prompt to render its target text independently within its assigned region.
This is enabled by constructing an isolated attention mask that implicitly regulates inter-region interactions, ensuring that each region focuses solely on its target text.
Specifically, we enforce that only the text and image tokens associated with a particular region (i.e., $\{p_i, r_i\}$) can attend to one another, both across and within modalities.
Let $m_{p_i} \in \{0,1\}^{L_T}$ and $m_{r_i} \in \{0,1\}^{L_I}$ be binary masks that activate only the token indices of the $i$-th textual prompt and its corresponding region, respectively. Here, $L_T$ is the total number of text tokens across all $n{+}1$ prompts, and $L_I$ is the number of image patch tokens (i.e., $h \times w$). By concatenating $m_{p_i}$ and $m_{r_i}$, we construct a joint mask $m_i \in \{0,1\}^{L_T + L_I}$ that activates all text and image tokens associated with the $i$-th region, while masking out the rest.
We then obtain the region-isolation attention mask $M_{\mathrm{isol}}$ by summing the outer products of these masks:
\begin{equation}
M_{\mathrm{isol}} = \sum_{i=1}^{n+1} m_i  \cdot m_i^{\top} \in \{0,1\}^{(L_T + L_I) \times (L_T + L_I)},
\label{eq:m_isol}
\end{equation}
where $\cdot$ denotes an outer product, resulting in a square mask.


However, while applying this mask allows each region to render its target text correctly, the resulting image fails to follow the global prompt. This occurs because $M_{\mathrm{isol}}$ eliminates all information flow between textual regions and the background, causing each region to be generated independently according to its own prompt. As a result, the final image may appear unnatural, as if multiple disjoint images were stitched together. Moreover, the background region, guided solely by the global prompt, may re-render text that is already rendered within the individual textual regions (left image in~\cref{fig:ablation}). In other words, this approach only divides the task but fails to conquer it.

\paragraph{Text-Focus Attention Mask.}





To address the limitations of $M_{\mathrm{isol}}$, \method{} expands the attendable regions in the mask (\ie areas with value 1), allowing directional information flow between the textual regions, the background region, the textual prompts, and the global prompt.

For the background region, we allow it to attend to the textual regions so that the background can connect naturally to the text areas and avoid redundant text generation within the background. The subscript arrow $r^c \to \{r_i\}$ indicates the direction of attention from query to key:
\begin{equation}
M_{r^c \to \{r_i\}} = m_{r^c} \cdot \mathbf{1}^\top_{L_I} \in \{0,1\}^{L_I \times L_I}.
\end{equation}
We also allow the background region to be attended by the textual prompts, enabling each prompt to incorporate surrounding visual context. This helps the prompts generate more natural and contextually appropriate descriptions, leading to more accurate text rendering:
\begin{equation}
M_{\{p_i\} \to r^c} = \mathbf{1}_{L_T} \cdot m_{r^c}^\top \in \{0,1\}^{L_T \times L_I}.
\end{equation}

For the global prompt, we apply a similar strategy: enabling it to attend to the textual regions and receive attention from the textual prompts. This promotes better coordination across the image and contributes to more natural and accurate text rendering (see Appendix~\ref{sec-b1:Text-Focus Attention Mask Design}):
\begin{equation}
\begin{split}
M_{p_g \to \{r_i\}} &= m_{p_g} \cdot \mathbf{1}^\top_{L_I} \in \{0,1\}^{L_T \times L_I}, \\
M_{\{p_i\} \to p_g} &= \mathbf{1}_{L_T} \cdot m_{p_g}^\top \in \{0,1\}^{L_T \times L_T}.
\end{split}
\end{equation}

Finally, the text-focus attention mask $M_{\mathrm{focus}}$ is constructed by combining $M_{\mathrm{isol}}$ with the four additional partial masks defined above:
\begin{equation}
M_{\mathrm{focus}} = M_{\mathrm{isol}} \lor
\begin{bmatrix}
M_{\{p_i\} \to p_g} & M_{\{p_i\} \to r^c} \lor M_{p_g \to \{r_i\}} \\
0 & M_{r^c \to \{r_i\}}
\end{bmatrix}.
\end{equation}

\paragraph{Context-Expansion Attention Mask.}

While $M_{\mathrm{focus}}$ enables accurate text generation within the textual regions and allows the background to naturally incorporate these regions, the textual regions themselves still attend solely to their own content.
In other words, each region remains blind to the background, potentially leading to an interior that looks visually isolated from its surroundings (see Appendix~\cref{fig:expn_mask_steps}).
To address this, we allow textual regions to attend to the background after a few denoising steps with $M_{\mathrm{focus}}$, once the target texts have been reasonably placed.
This enables bidirectional information flow between the regions and the background:
\begin{equation}
\begin{split}
M_{\{r_i\} \to r^c} &= \mathbf{1}_{L_I} \cdot m_{r^c}^\top \in \{0,1\}^{L_I \times L_I}, \\
M_{\mathrm{expn}} &= M_{\mathrm{focus}} \lor
\begin{bmatrix}
0 & 0 \\
0 & M_{\{r_i\} \to r^c}
\end{bmatrix}.
\end{split}
\end{equation}

\subsection{Localized Noise Initialization}
\label{Localized Noise Initialization}

We observe that generating text over an entire region-sized image is often easier than placing the same text within that region in the context of a full-sized image.
Motivated by this, we propose a initialization strategy (\cref{fig:mask}a) that performs light denoising on region-specific noise patches, allowing each region to receive early guidance before global denoising begins.
We first sample initial noise for the full image, $z_T \sim \mathcal{N}(0, I)$, where $z_T \in \mathbb{R}^{c \times h \times w}$.
From $z_T$, we extract a set of localized noise patches $\{z_{T_i} \in \mathbb{R}^{c \times h_i \times w_i}\}_{i=1}^n$, each corresponding to a textual region $\{r_i\}_{i=1}^n$.
Each patch $z_{T_i}$ is then independently denoised for a very small number of steps using its associated textual prompt, yielding a lightly refined latent $z_{T'_i}$, where $T' = T - T_\mathrm{init}$.
These refined patches are blended back into their original locations within $z_T$ using a weighting factor $\alpha$, resulting in the updated global latent $z_{T'}$. Specifically, for each region $r_i$:
\begin{equation}
z_{T'}[r_i]=\alpha \cdot z_{T_i} + (1 - \alpha) \cdot z_{T'_i}.
\end{equation}

After this initialization for $T_\mathrm{init}$ steps, we proceed with the remaining $T'$ denoising steps using the updated latent $z_{T'}$.
This simple initialization approach improves region-text spatial alignment and text rendering accuracy while remaining computationally efficient, as the total region area $\sum_i h_i w_i$ is smaller than the full image area $h w$.

\subsection{\boldmethod{} Pipeline}
\label{pipeline}

\method{} divides the denoising process into four sequential stages: (1) Localized Noise Initialization for $T_\mathrm{init}$ steps, (2) Text-Focus denoising using the attention mask $M_{\mathrm{focus}}$ for $T_\mathrm{focus}$ steps, (3) Context-Expansion denoising using the attention mask $M_{\mathrm{expn}}$ for $T_\mathrm{expn}$ steps, and (4) standard denoising without attention masking control, using only the global prompt for the remaining steps.
The full pseudocode of the \method{} pipeline is provided in~\cref{alg:dctext}.

\section{Experiments}
\label{Experiments}

\subsection{Experiment Setting}
\label{Experiment Setting}

\paragraph{Benchmark.}
We evaluate our method on four datasets: ChineseDrawText~\cite{ma2023glyphdraw}, DrawTextCreative~\cite{liu2022character}, TMDBEval500~\cite{chen2023textdiffuser}, and CVTG-Style~\cite{du2025textcrafter}. The first three datasets mainly contain single-sentence prompts, while CVTG-Style includes prompts with 2–5 rendering sentences.
These datasets cover diverse text rendering scenarios, from artistic typography to real-world scenes and stylized text.

\paragraph{Metric.}

We evaluate text rendering accuracy using sentence-level accuracy (Acc.) and Normalized Edit Distance (NED)~\cite{tuo2023anytext}, with PP-OCRv4~\cite{ppocr} for single-sentence prompts and GPT-4o~\cite{hurst2024gpt} for multi-sentence prompts to handle sentence-level recognition. Image quality is assessed using Q-Align~\cite{wu2023q} for Quality and Aesthetic Scores, and CLIP Score~\cite{clip} for prompt–image alignment. Latency per image generation is measured for efficiency, and a user study is conducted to assess preference.

\begin{table}[t]\centering

\footnotesize
\begin{adjustbox}{max width=\columnwidth}
\begin{tabular}{l|rr|rrr|rr}
\toprule
Method & Acc. & NED & CLIP & Qual. & Aesth. & Steps & Latency \\
\midrule
Flux & 0.266 & 0.579 & 0.343 & 4.657 & 3.851 & 24 & 13.89 \\
AMO & 0.274 & 0.569 & 0.342 & 4.658 & 3.863 & 28 & 25.93 \\
TC & 0.329 & 0.722 & \textbf{0.350} & 4.704 & \textbf{3.911} & 30 & 36.89 \\
\textbf{Ours} & \textbf{0.387} & \textbf{0.751} & 0.349 & \textbf{4.737} & 3.904 & 24 & 16.60 \\
\bottomrule
\end{tabular}

\end{adjustbox}
\caption{\textbf{Quantitative comparison on single sentence.} Results are averaged over three single-sentence datasets (ChineseDrawText~\cite{ma2023glyphdraw}, DrawTextCreative~\cite{liu2022character}, TMDBEval500~\cite{chen2023textdiffuser}). Steps and latency indicate denoising steps and total generation time per image (seconds, measured on an L40 GPU).}
\label{tab:single_comparison}

\end{table}

\begin{table}[!t]\centering

\footnotesize
\begin{adjustbox}{max width=\columnwidth}
\begin{tabular}{lr|rr|rrr|rr}
\toprule
$n$ & Method & Acc. & NED & CLIP & Qual. & Aesth. & Latency \\
\midrule\midrule
\multirow{4}{*}{$5$}
&Flux &0.366 &0.608 &0.336 &4.591 &3.165 &13.91 \\
&AMO &0.432 &0.660 &0.335 &4.652 &3.218 &26.02 \\
&TC &0.685 &0.859 &\textbf{0.349} &4.659 &3.506 &40.53 \\
&\textbf{Ours} &\textbf{0.693} &\textbf{0.860} &0.343 &\textbf{4.697} &\textbf{3.569} &19.26 \\
\midrule
\multirow{4}{*}{$4$}
&Flux &0.389 &0.628 &0.338 &4.707 &3.421 &13.93 \\
&AMO &0.488 &0.690 &0.337 &4.709 &3.518 &25.97 \\
&TC &0.693 &0.867 &0.352 &4.665 &3.488 &39.60 \\
&\textbf{Ours} &\textbf{0.760} &\textbf{0.892} &\textbf{0.353} &\textbf{4.745} &\textbf{3.659} &18.14 \\
\midrule
\multirow{4}{*}{$3$}
&Flux &0.508 &0.715 &0.345 &4.662 &3.377 &13.88 \\
&AMO &0.575 &0.750 &0.343 &4.689 &3.489 &25.99 \\
&TC &0.722 &0.880 &\textbf{0.351} &4.659 &3.571 &39.01 \\
&\textbf{Ours} &\textbf{0.768} &\textbf{0.906} &\textbf{0.351} &\textbf{4.735} &\textbf{3.709} &16.96 \\
\midrule
\multirow{4}{*}{$2$}
&Flux &0.608 &0.809 &0.344 &4.675 &3.471 &13.88 \\
&AMO &0.642 &0.826 &0.341 &4.654 &3.584 &25.91 \\
&TC &0.758 &0.919 &0.346 &4.697 &3.616 &38.22 \\
&\textbf{Ours} &\textbf{0.792} &\textbf{0.923} &\textbf{0.347} &\textbf{4.791} &\textbf{3.726} &15.66 \\
\bottomrule
\end{tabular}
\end{adjustbox}

\caption{\textbf{Quantitative comparison on multiple sentences.} Comparison results on the CVTG-Style~\cite{du2025textcrafter} dataset across different numbers of sentences ($n$) to be rendered.}
\label{tab:multi_comparison}

\end{table}


\paragraph{Baselines.}
We compare our method against tuning-free visual text generation methods based on Flux.1-dev~\cite{flux1-dev}:
Flux, the base model;
AMO Sampler (AMO)~\cite{hu2025amosamplerenhancingtext}, a sampler that adaptively corrects sampling trajectories using attention weights;
TextCrafter (TC)~\cite{du2025textcrafter}, explicitly isolating text regions for denoising and enhances attention for clearer rendering.
All baselines use official implementations with default settings.
Further comparisons with training-based methods and tuning-free methods based on Stable Diffusion 3.5 are provided in Appendix~\ref{sec-a3:Comparison with Training-based Methods} and~\ref{sec-c2:Stable Diffusion 3.5}.

\paragraph{Implementation Details.}

We adopt Flux.1-dev~\cite{flux1-dev} as our base model and generate all images at $1024\times1024$ resolution with 24 denoising steps.
For each sentence in a prompt, we construct a textual prompt and region, using GPT-4o~\cite{hurst2024gpt} for single-sentence cases and adopting CVTG-Style prompts while sharing regions from TextCrafter~\cite{du2025textcrafter} for multi-sentence cases.
We set $(T_{\mathrm{init}}, T_{\mathrm{focus}}, T_{\mathrm{expn}})$ to (1,2,2) for single-sentence generation and (2,3,2) for multi-sentence generation; more details are in Appendix~\ref{sec-e1:Implementation}.

\begin{figure}[!t]
  \centering
    \includegraphics[width=0.95\columnwidth]{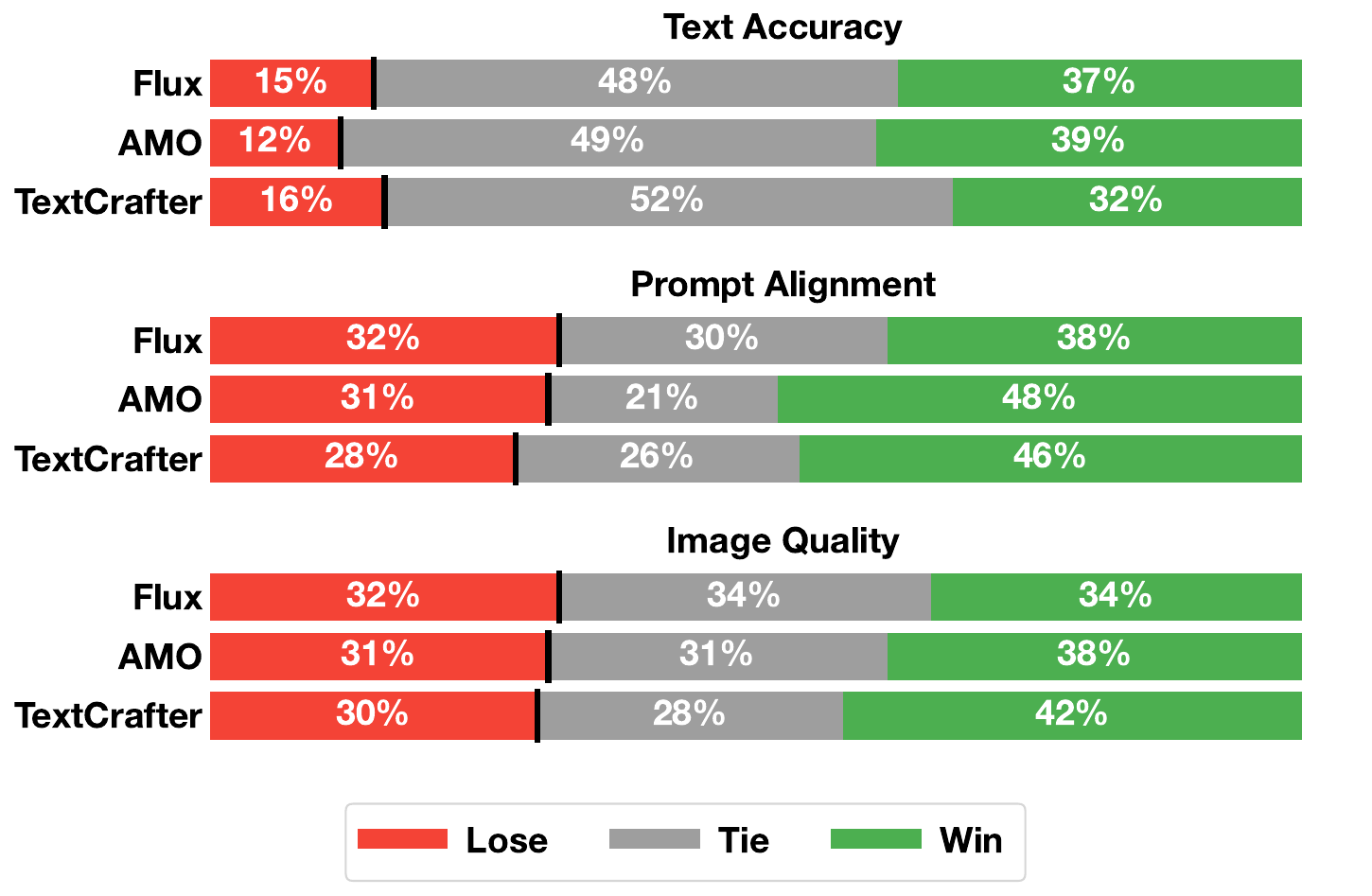} 
    \caption{\textbf{Human Evaluation.} User preference on text accuracy, prompt–image alignment, and overall image quality. Green bars indicate cases where our method is preferred.}
    \label{fig:human}
\end{figure}

\subsection{Qualitative Results}

\cref{fig:single} compares \method{} with all tuning-free baselines on datasets containing single rendering sentences.
In columns 1 and 4, baselines omit the sentence or produce only partial outputs, while our method generates the full text.
In columns 2, 5, and 6, baselines generate incorrect or repeated text, whereas \method{} produces the correct text as prompted.
Notably, TextCrafter often produces text regions visually detached from the rest of the image (columns 3–5, \eg with awkward white backgrounds), resulting in unnatural compositions.
In contrast, \method{} integrates text seamlessly into the scene, maintaining overall visual coherence.


\cref{fig:multi} further compares our method with TextCrafter on multi-sentence generation. TextCrafter frequently fails to generate text within the designated areas or renders unintended text outside them (rows 1–3).
As the number of regions increases (row 4), it often generates phrases in incorrect regions (\eg \textit{``OUTDOOR SEATING OPEN''}) or repeats the same phrase across different regions (\eg \textit{``Family Meals Available''}).
This stems from its global attention operating over the entire image, which introduces cross-region interference even when denoising is applied independently per region.
In contrast, with Localized Noise Initialization and the Text-Focus Attention Mask, our method enforces region-focused attention control, accurately rendering the text in each correct region while preventing cross-region interference. At the same time, background region remains text-free and blend naturally with the surrounding scene. Additional samples are provided in Appendix~\ref{sec-a1:Qualitative Samples}.

\subsection{Quantitative Results}
\cref{tab:single_comparison} presents the results on three single-sentence datasets. In terms of text accuracy, \method{} achieves 45\% and 30\% improvements over Flux in sentence-level accuracy and NED, respectively, outperforming all baselines. 
Notably, our method maintains high aesthetic and quality scores, showing that these improvements do not compromise generative priors of the base model.
The high CLIP score further suggests that the model continues to follow the global prompt well, even when conditioned on decomposed multiple prompts.
Remarkably, these gains are achieved with the fewest denoising steps and the lowest latency, requiring only about 20\% additional latency compared to Flux. Latency is measured over the entire pipeline, including Localized Noise Initialization (1 step) and 23 denoising steps.

\cref{tab:multi_comparison} compares \method{} against all baselines on the multi-sentence dataset across varying numbers of regions.
As in~\cref{tab:single_comparison}, our method consistently achieves the highest text accuracy without compromising image quality across all region counts.
These results demonstrate the effectiveness of our divide-and-conquer strategy, which decomposes text by region and employs scheduled attention masking to ensure accurate and coherent text generation within the overall image context.
Although latency increases slightly as the number of regions grows, it remains lower than all baselines except Flux, underscoring efficiency of \method{}. More comprehensive results are available in Appendix~\ref{sec-a2:Quantitative Results}.

To further assess user preferences, we conduct a human evaluation across three datasets, assessing text accuracy, alignment between the image and the global prompt, and overall image quality, following the same protocol as~\cref{tab:single_comparison}. A total of 30 evaluators perform pairwise comparisons between \method{} and each baseline over 1,323 image pairs (see Appendix~\ref{sec-e2:Human Evaluation}). The results, presented in~\cref{fig:human}, show that our method is consistently favored across all criteria, with particularly significant improvements in text accuracy.

\subsection{Ablation Study}
\label{Ablation Study}
\begin{figure}[!t]
  \centering
    \includegraphics[width=0.95\columnwidth]{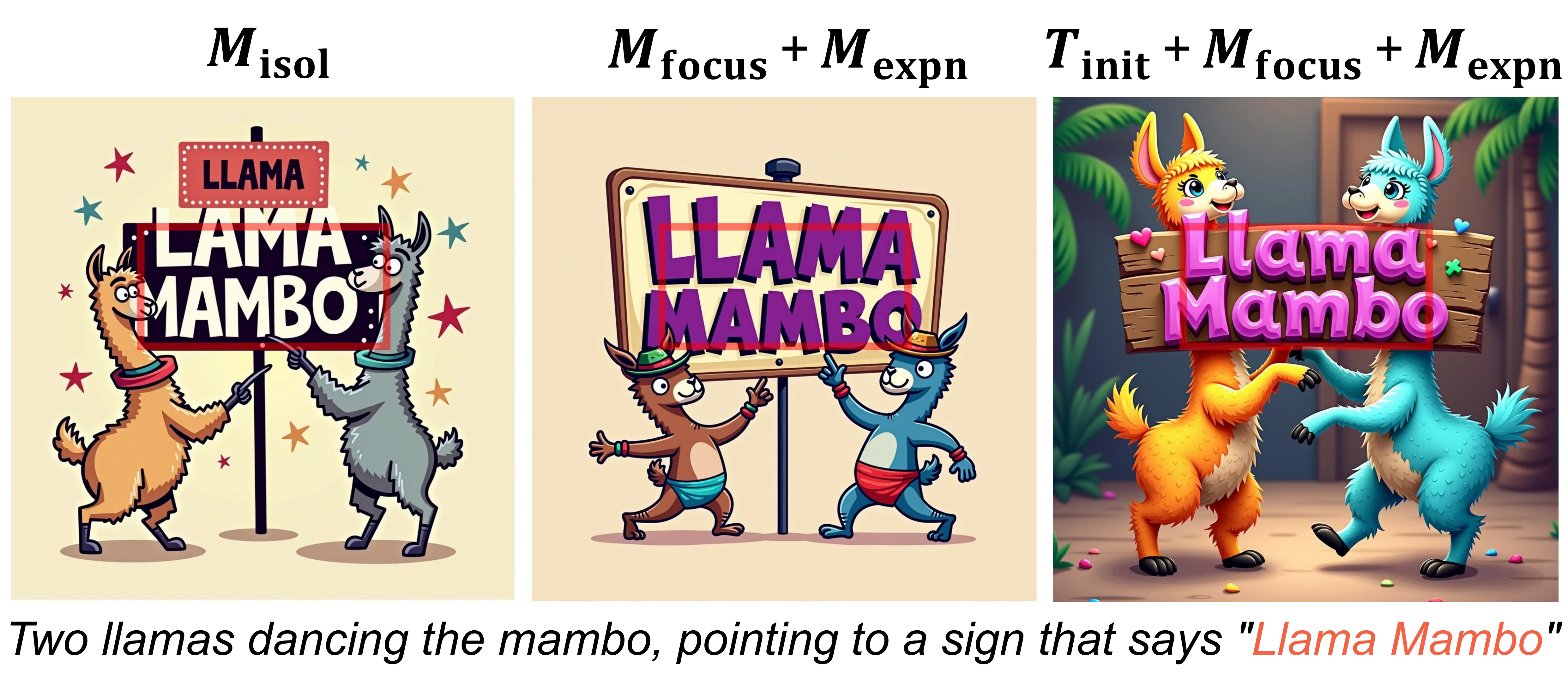} 
    \caption{\textbf{Qualitative results of ablation study.} $M_{\mathrm{isol}}$ causes redundant text and harsh boundary, while $M_{\mathrm{focus}} + M_{\mathrm{expn}}$ yields accurate text and smooth transition. Localized Noise Initialization further enhances text-region alignment}
    \label{fig:ablation}
\end{figure}

\cref{tab:ablation} and~\cref{fig:ablation} present ablation results, evaluating each component of \method{}. 
To assess the effectiveness of our proposed masks, we compare the combination of $M_{\mathrm{focus}}$ and $M_{\mathrm{expn}}$ against using only $M_{\mathrm{isol}}$ during the $T_\mathrm{focus} + T_\mathrm{expn}$ steps.
As shown in~\cref{fig:ablation} (left) and the first row of~\cref{tab:ablation}, using $M_{\mathrm{isol}}$ results in redundant text rendering and lower text accuracy (Acc., NED). This is due to the textual region and background independently attending to separate prompts—a textual prompt and a global prompt. In addition, the restricted information flow enforced by $M_{\mathrm{isol}}$ creates a hard boundary around the region, preventing smooth transitions between the textual region and the background.
In contrast, our two-stage masks ($M_{\mathrm{focus}} + M_{\mathrm{expn}}$) first focus attention on the textual region to prevent redundancy, and then gradually expand it to the background, enabling seamless transitions. 
This improves both text accuracy and overall image quality by achieving precise attention control between region and background.
Finally, denoising from an latent obtained via Localized Noise Initialization (\ie full \method{} pipeline) provides region-specific guidance, achieving tighter alignment between rendered text and its region (right image in~\cref{fig:ablation}) and further boosting text accuracy (last row in~\cref{tab:ablation}).
The sensitivity analysis of each component can be found in Appendix~\ref{sec-b2:Text-Focus Denoising Steps}--\ref{sec-b4:Localized Noise Initialization Steps}.

\begin{table}[!t]\centering

\footnotesize
\begin{adjustbox}{max width=\columnwidth}
\begin{tabular}{l|rr|rrr}
\toprule
Method & Acc. & NED & CLIP & Qual. & Aesth. \\
\midrule
$M_{\mathrm{isol}}$ &0.222 &0.640 &0.345 &4.564 &3.683 \\
$M_{\mathrm{focus}}$ + $M_{\mathrm{expn}}$ &0.306 &0.700 &0.346 &4.569 &3.700 \\
$T_{\mathrm{init}}$ + $M_{\mathrm{focus}}$ + $M_{\mathrm{expn}}$ &\textbf{0.387} &\textbf{0.751} &\textbf{0.349} &\textbf{4.737} &\textbf{3.904} \\
\bottomrule
\end{tabular}
\end{adjustbox}

\caption{\textbf{Ablation study on \boldmethod{} components.} $M_{\mathrm{isol}}$ denotes denoising using only the isolation attention mask. $M_{\mathrm{focus}}$ + $M_{\mathrm{expn}}$ uses our attention masks during two-stage denoising. $T_{\mathrm{init}}$ + $M_{\mathrm{focus}}$ + $M_{\mathrm{expn}}$ further incorporates Localized Noise Initialization, forming our full pipeline.}
\label{tab:ablation}

\end{table}


\section{Conclusion}
\label{Conclusion}


We present \method{}, a training-free method that adopts a divide-and-conquer strategy for accurate and coherent visual text generation. By decomposing target texts and guiding generation with scheduled attention masking, \method{} improves text accuracy with the lowest latency among all compared methods, without degrading image quality.

\paragraph{Acknowledgement}
This work was supported by the National Research Foundation of Korea (NRF) grant funded by the Korea government (MSIT) [No. 2022R1A3B1077720; 2022R1A5A7083908], Institute of Information \& Communications Technology Planning \& Evaluation (IITP) grant funded by the Korea government (MSIT) [RS-2022-II220959; No.RS-2021-II211343, Artificial Intelligence Graduate School Program (Seoul National University)], the BK21 FOUR program of the Education and Research Program for Future ICT Pioneers, Seoul National University in 2025.

{
    \small
    \bibliographystyle{ieeenat_fullname}
    \bibliography{main}
}

\clearpage
\appendix
\setcounter{section}{0}
\renewcommand\thesection{\Alph{section}}
\setcounter{table}{0}
\renewcommand{\thetable}{S\arabic{table}}
\setcounter{figure}{0}
\renewcommand{\thefigure}{S\arabic{figure}}

\section*{\Large{Appendix}}

\section{Additional Results}
\label{sec-a:Additional Results}

\subsection{Qualitative Samples}
\label{sec-a1:Qualitative Samples}
\begin{figure*}[!t]
  \centering
    \includegraphics[width=0.95\textwidth]{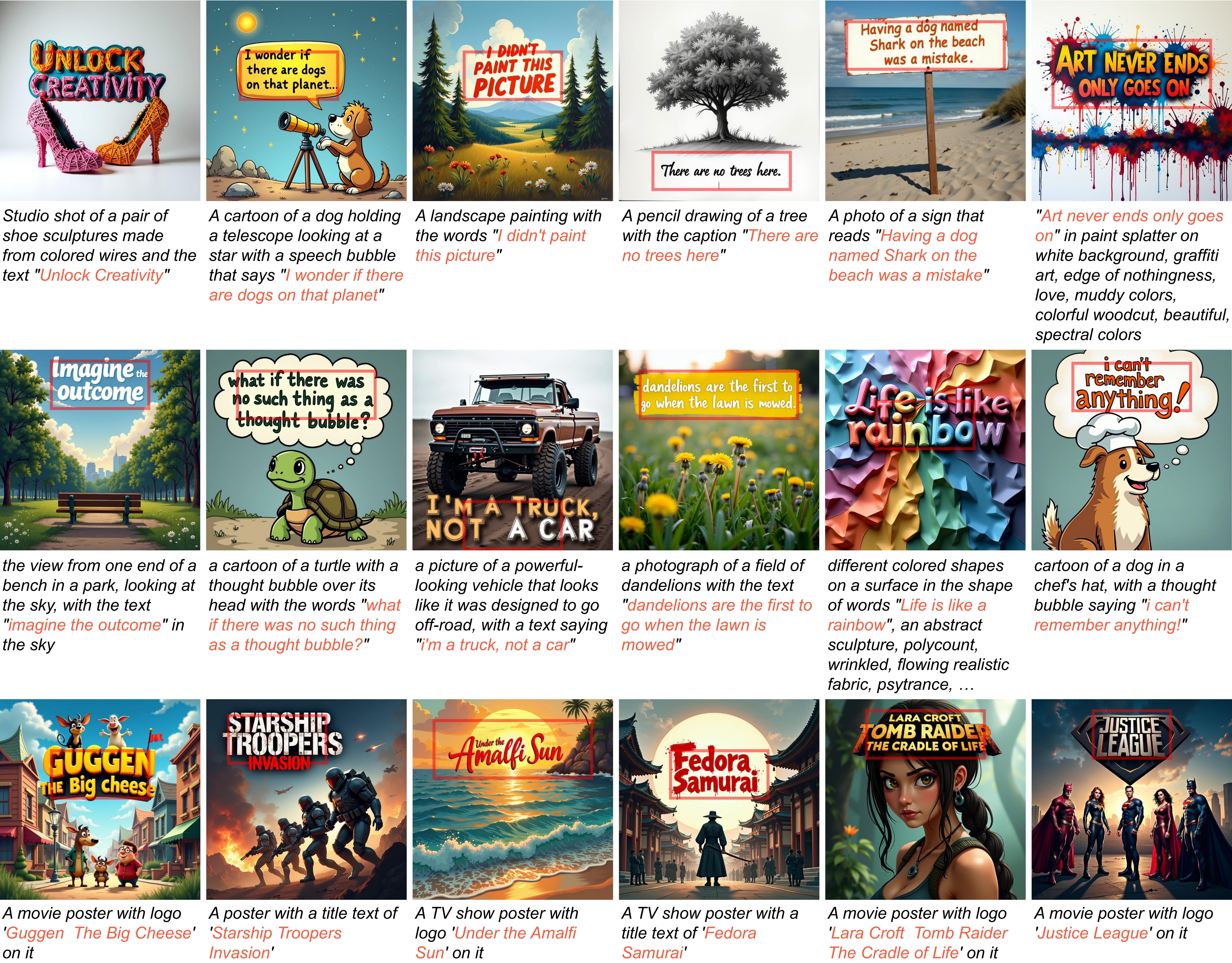}
    
    \caption{\textbf{Qualitative samples on single sentence.} Prompts, including the sentence to be rendered (highlighted in red), are shown below each image. Corresponding textual regions are indicated with red boxes.}
    \label{fig:qulity_single}
\end{figure*}

\begin{figure*}[!htbp]
  \centering
    \includegraphics[width=0.9\textwidth]{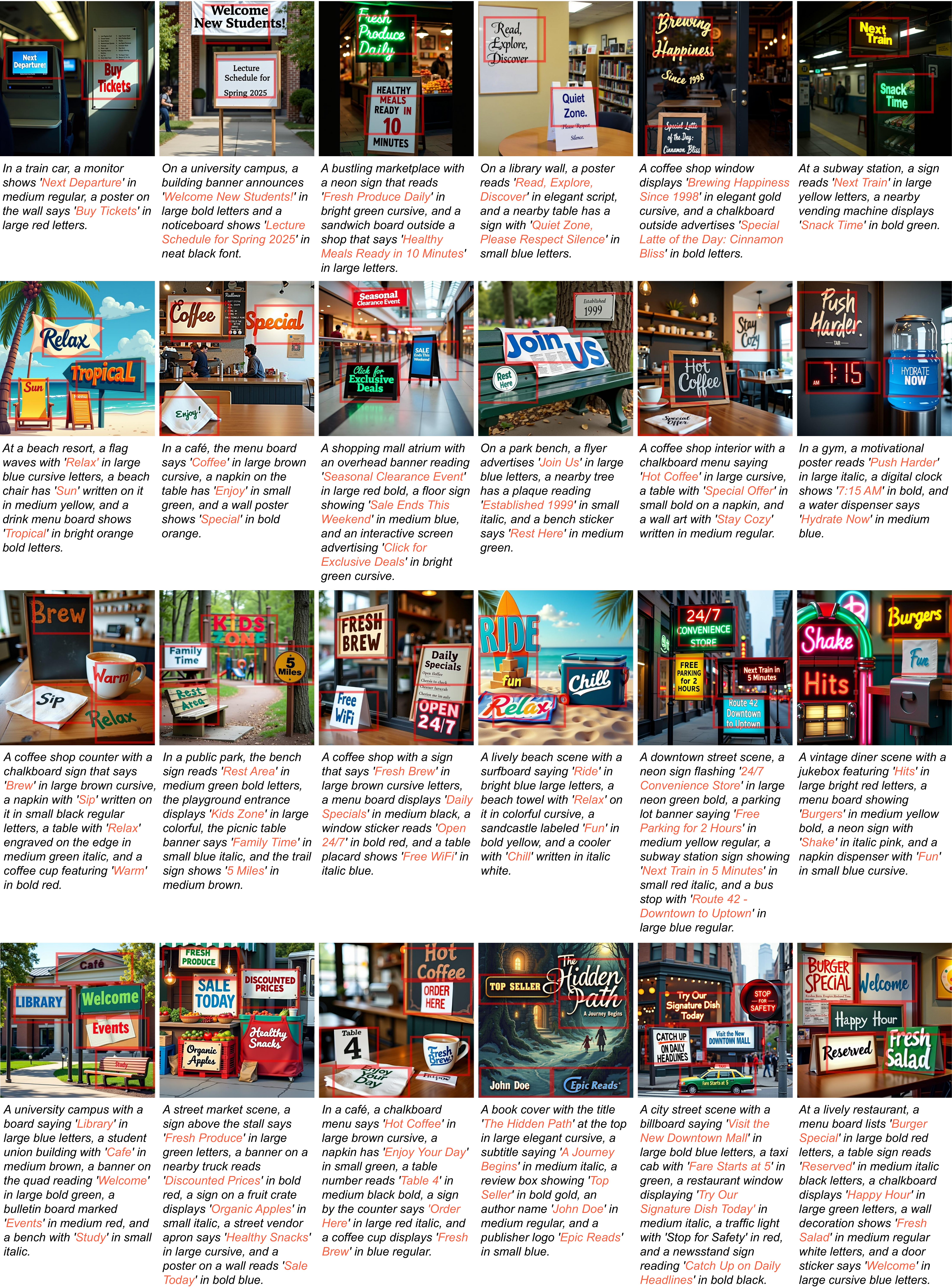}
    \caption{\textbf{Qualitative samples on multiple sentences.} Prompts, including the sentences to be rendered (highlighted in red), are shown below each image. Corresponding textual regions are indicated with red boxes.}
    \label{fig:quality_multi}
\end{figure*}

\cref{fig:qulity_single} and~\cref{fig:quality_multi} present additional qualitative samples generated by \method{}. As shown in the results, our method consistently produces accurate text and high-quality images across a wide range of themes, including illustrations, real scenes, posters, artistic styles, and animations. Notably, these results are achieved without any additional training, relying solely on scheduled attention masking at inference time.
In particular,~\cref{fig:quality_multi} demonstrates that the textual regions faithfully reflect the target sentences, even when multiple texts need to be rendered. At the same time, the overall image remains thematically aligned with the global prompt.
This is enabled by \method{}’s attention masks ($M_{\mathrm{focus}}$, $M_{\mathrm{expn}}$), which effectively regulate the information flow between the target regions and the background.

\subsection{Quantitative Results}
\label{sec-a2:Quantitative Results}
\cref{tab:full_comparison} provides a comprehensive comparison across all datasets and baselines. To support broader evaluation, we additionally report baseline results obtained using the same number of denoising steps as our method, indicated by †.
For text accuracy metrics (Acc. and NED), \method{} consistently outperforms all baselines across all datasets. This underscores the effectiveness of our divide-and-conquer strategy, which generates text segments rather than entire sentences at once, resulting in more reliable text generation. 
In terms of overall image quality (Qual. and Aesth.), our method also achieves high scores on most datasets, indicating that improvements in text accuracy do not come at the expense of image quality.
In addition, when baselines are evaluated under the same reduced number of denoising steps, their performance typically declines across most metrics. Although this setting reduces inference latency, our method still maintains the lowest latency, with strong performance.

\begin{table*}[!t]\centering
\begin{adjustbox}{max width=0.88\textwidth}

\begin{tabular}{l|r|r|rr|rrr|rrr}\toprule
Dataset &$n$ &Method &Acc. &NED &CLIP &Qual. &Aesth. &Steps &Latency (sec.) \\
\midrule\midrule

\multirow{6}{*}{CreativeDrawText} &\multirow{6}{*}{1} &Flux &0.257 &0.544 &0.339 &4.695 &3.718 &24 &13.89 \\
& &AMO &0.261 &0.559 &0.339 &4.691 &3.693 &28 &25.93   \\
& &AMO† &0.193 &0.524 &0.338 &4.697 &3.707 &24 &21.71   \\
& &TextCrafter &0.330 &0.764 &\textbf{0.346} &4.726 &\textbf{3.767} &30 &36.89 \\
& &TextCrafter† &0.330 &0.750 &\textbf{0.346} &4.727 &3.722 &24 &28.61 \\
& &\textbf{DCText (Ours)} &\textbf{0.427} &\textbf{0.809} &0.345 &\textbf{4.775} &3.761 &24 &16.60 \\

\midrule
\multirow{6}{*}{DrawTextCreative} &\multirow{6}{*}{1} &Flux &0.223 &0.525 &0.351 &4.657 &3.774 &24 &13.81 \\
& &AMO &0.234 &0.499 &0.351 &4.670 &3.831 &28 &25.99 \\
& &AMO† &0.234 &0.516 &0.350 &4.624 &3.755 &24 &21.76 \\
& &TextCrafter &0.286 &0.645 &0.351 &4.698 &3.848 &30 &36.92 \\
& &TextCrafter† &0.286 &0.664 &\textbf{0.353} &4.691 &3.855 &24 &28.61 \\
& &\textbf{DCText (Ours)} &\textbf{0.337} &\textbf{0.675} &0.350 &\textbf{4.765} &\textbf{3.934} &24 &16.61 \\

\midrule
\multirow{6}{*}{TMDBEval500} &\multirow{6}{*}{1} &Flux &0.318 &0.667 &0.340 &4.618 &4.061 &24 &13.88 \\
& &AMO &0.326 &0.648 &0.336 &4.614 &4.064 &28 &25.89 \\
& &AMO† &0.340 &0.665 &0.338 &4.607 &4.041 &24 &21.74 \\
& &TextCrafter &0.372 &0.758 &\textbf{0.352} &\textbf{4.687} &\textbf{4.119} &30 &36.91 \\
& &TextCrafter† &0.358 &0.744 &0.351 &4.661 &4.074 &24 &28.63 \\
& &\textbf{DCText (Ours)} &\textbf{0.396} &\textbf{0.768} &0.351 &4.670 &4.018 &24 &16.56 \\

\midrule
\multirow{24}{*}{CVTG-Style}

&\multirow{6}{*}{2} &Flux &0.608 &0.809 &0.344 &4.675 &3.471 &24 &13.88 \\
& &AMO &0.642 &0.826 &0.341 &4.654 &3.584 &28 &25.91 \\
& &AMO† &0.622 &0.820 &0.341 &4.664 &3.580 &24 &21.98 \\
& &TextCrafter &0.758 &0.919 &0.346 &4.697 &3.616 &30 &38.22 \\
& &TextCrafter† &0.745 &0.923 &0.348 &4.688 &3.584 &24 &28.97 \\
& &\textbf{DCText (Ours)} &\textbf{0.792} &\textbf{0.923} &\textbf{0.347} &\textbf{4.791} &\textbf{3.726} &24 &15.66 \\
\cmidrule(lr){2-10}

&\multirow{6}{*}{3} &Flux &0.508 &0.715 &0.345 &4.662 &3.377 &24 &13.88 \\
& &AMO &0.575 &0.750 &0.343 &4.689 &3.489 &28 &25.99 \\
& &AMO† &0.540 &0.741 &0.342 &4.679 &3.480 &24 &21.86 \\
& &TextCrafter &0.722 &0.880 &\textbf{0.351} &4.659 &3.571 &30 &39.01 \\
& &TextCrafter† &0.710 &0.882 &0.350 &4.670 &3.595 &24 &29.47 \\
& &\textbf{DCText (Ours)} &\textbf{0.768} &\textbf{0.906} &\textbf{0.351} &\textbf{4.735} &\textbf{3.709} &24 &16.96 \\
\cmidrule(lr){2-10}

&\multirow{6}{*}{4} &Flux &0.389 &0.628 &0.338 &4.707 &3.421 &24 &13.93 \\
& &AMO &0.488 &0.690 &0.337 &4.709 &3.518 &28 &25.97 \\
& &AMO† &0.469 &0.689 &0.337 &4.709 &3.515 &24 &21.97 \\
& &TextCrafter &0.693 &0.867 &0.352 &4.665 &3.488 &30 &39.60 \\
& &TextCrafter† &0.722 &0.877 &0.354 &4.697 &3.530 &24 &30.07 \\
& &\textbf{DCText (Ours)} &\textbf{0.760} &\textbf{0.892} &\textbf{0.353} &\textbf{4.745} &\textbf{3.659} &24 &18.14 \\
\cmidrule(lr){2-10}

&\multirow{6}{*}{5} &Flux &0.366 &0.608 &0.336 &4.591 &3.165 &24 &13.91 \\
& &AMO &0.432 &0.660 &0.335 &4.652 &3.218 &28 &26.02 \\
& &AMO† &0.402 &0.636 &0.336 &4.618 &3.190 &24 &21.94 \\
& &TextCrafter &0.685 &0.859 &\textbf{0.349} &4.659 &3.506 &30 &40.53 \\
& &TextCrafter† &0.661 &0.848 &0.343 &4.562 &3.396 &24 &30.78 \\
& &\textbf{DCText (Ours)} &\textbf{0.693} &\textbf{0.860} &0.343 &\textbf{4.697} &\textbf{3.569} &24 &19.26 \\

\midrule\midrule
\multirow{6}{*}{Average} &\multirow{6}{*}{-} &Flux &0.381 &0.642 &0.342 &4.658 &3.570 &24 &13.88 \\
& &AMO &0.423 &0.662 &0.340 &4.668 &3.628 &28 &25.96 \\
& &AMO† &0.400 &0.656 &0.340 &4.657 &3.610 &24 &21.84 \\
& &TextCrafter &0.549 &0.813 &\textbf{0.350} &4.684 &3.702 &30 &38.30 \\
& &TextCrafter† &0.545 &0.813 &0.349 &4.671 &3.679 &24 &29.31 \\
& &\textbf{DCText (Ours)} &\textbf{0.596} &\textbf{0.833} &0.349 &\textbf{4.740} &\textbf{3.768} &24 &17.11 \\
\bottomrule

\end{tabular}
\end{adjustbox}

\caption{\textbf{Full quantitative comparison.} Comparison results with baselines on four datasets: ChineseDrawText~\cite{ma2023glyphdraw}, DrawTextCreative~\cite{liu2022character}, TMDBEval500~\cite{chen2023textdiffuser}, and CVTG-Style~\cite{du2025textcrafter}. † indicates methods that use the same number of denoising steps as \method{}.}
\label{tab:full_comparison}
\end{table*}

\subsection{Comparison with Training-based Methods}
\label{sec-a3:Comparison with Training-based Methods}
\begin{figure*}[!t]
  \centering
    \includegraphics[width=0.923\textwidth]{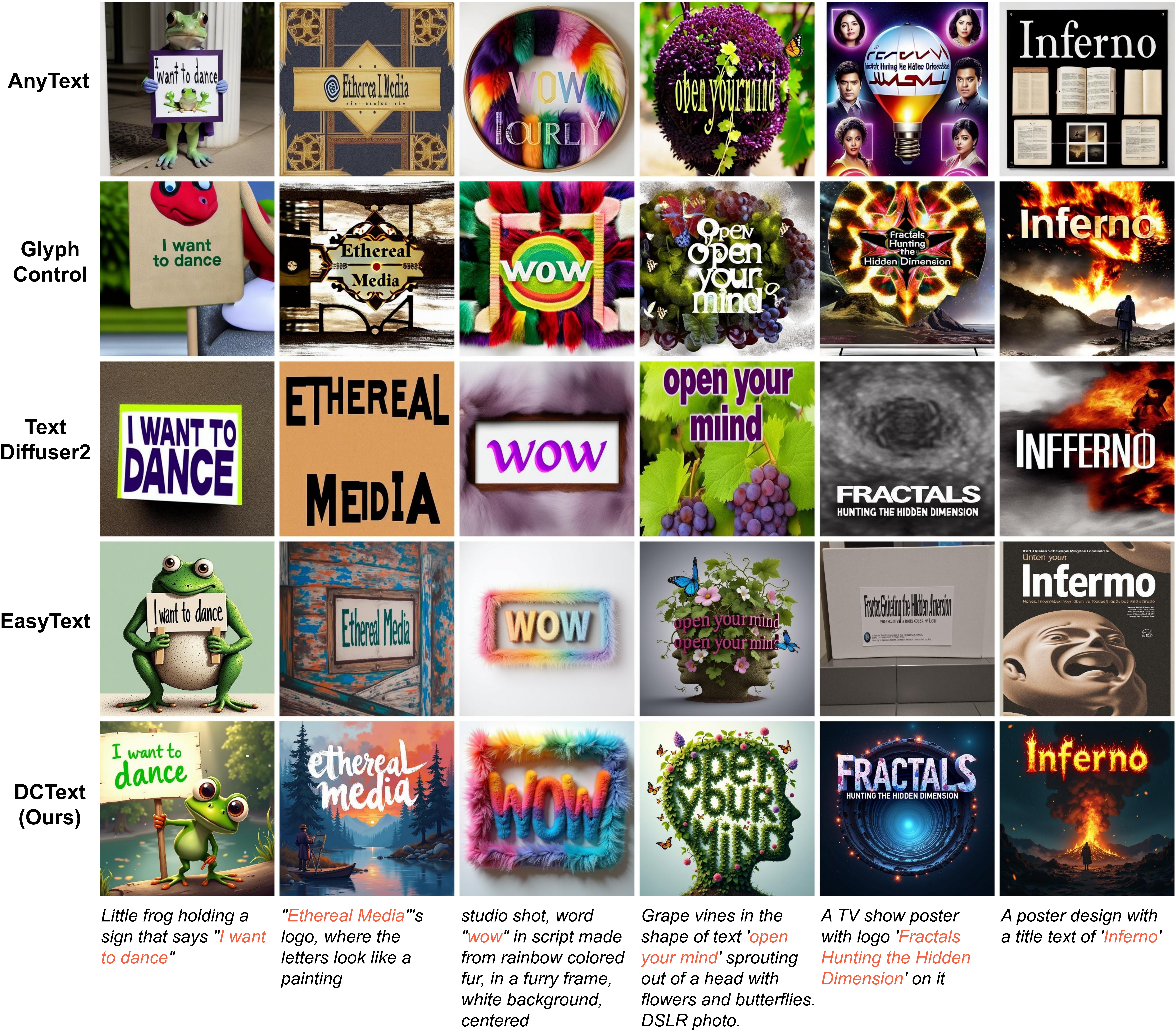}
    \caption{\textbf{Qualitative comparison between training-based baselines.}}
    \label{fig:train_single}
\end{figure*}

In the main paper, we compare \method{} with other approaches that, like ours, leverage a pre-trained text-to-image model without additional training.
We further compare our method against training-based approaches, including AnyText~\cite{tuo2023anytext}, GlyphControl~\cite{yang2023glyphcontrol}, TextDiffuser2~\cite{chen2024textdiffuser}, and EasyText~\cite{lu2025easytext}. \cref{tab:train_single} presents a quantitative comparison on the single-sentence datasets. In terms of text accuracy (Acc. and NED), training-based baselines—trained on large-scale text-centric datasets (AnyWord-3M~\cite{tuo2023anytext}, LAION-Glyph~\cite{yang2023glyphcontrol}, MARIO-10M~\cite{chen2023textdiffuser}, and EasyText-1M~\cite{lu2025easytext}) with glyph-level conditioning—generally achieve higher scores.
However, this comes at the cost of overall image quality. As reflected by their low aesthetic scores (Aesth.)—and as visually confirmed in~\cref{fig:train_single}—training-based methods lack stylistic diversity and fail to produce artistic text. For instance, in columns 3 and 4, where the prompts specify rendering text with fur and vines, these methods generate plain, generic text instead of following the intended styles.
In the more challenging multi-sentence setting, training-based approaches also struggle. As shown in~\cref{tab:train_multi}, their text accuracy degrades significantly as the number of sentences ($n$) increases. In contrast, \method{} maintains consistent performance and outperforms all baselines across different sentence counts.

\subsection{Comparison to Regional-Prompting}
\label{sec-a4:Comparison to Regional-Prompting}
\begin{figure*}[!t]
  \centering
    \includegraphics[width=0.8\textwidth]{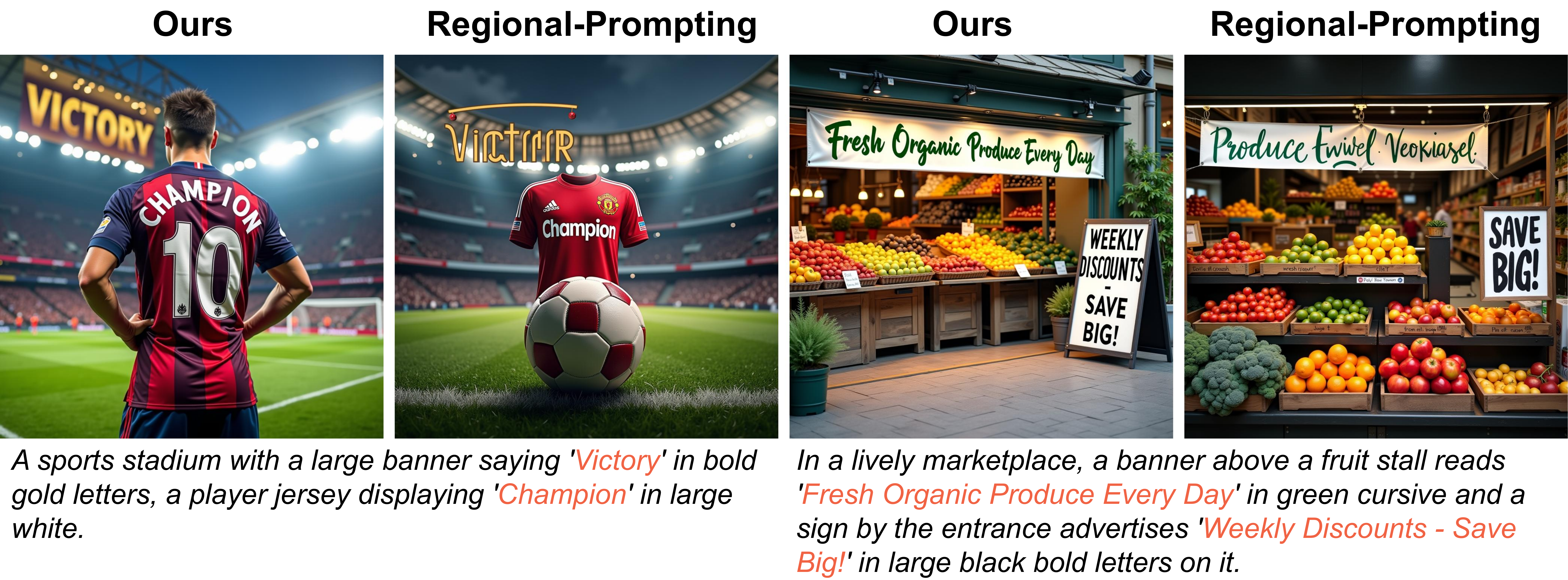}
    \caption{\textbf{Comparison to Regional-Prompting~\cite{chen2024training}} Comparison of generation results with another attention control method that relies solely on the Region-Isolation Attention Mask ($M_{\mathrm{isol}}$). For a fair comparison, we set $T_{\mathrm{init}}=0$.}
    \label{fig:comparison_rp}
\end{figure*}

To highlight \method{}’s performance in visual text generation, we compare it with Regional-Prompting~\cite{chen2024training}, a method that relies solely on the Region-Isolation Attention Mask ($M_{\mathrm{isol}}$) for inference-time attention control.
As discussed in~\cref{Attention Mask Control}, the exclusive use of $M_{\mathrm{isol}}$ often results in redundant text rendering and unnatural regional artifacts (see Fig. 6, left). Regional-Prompting addresses this by replacing the global prompt with a background-only prompt, removing all content information, and additionally performs a separate denoising process with the original global prompt, spatially blending the two resulting latents.
However, this approach not only doubles the number of function evaluations (NFEs), but also remains less effective for visual text generation.
As shown in~\cref{fig:comparison_rp}, Regional-Prompting often produces illegible or semantically meaningless text. This occurs because the fine-detailed visual features required for faithful text rendering are diluted during latent blending.
In contrast, \method{} generates text that is both accurate and natural, while also requiring only 15.66 seconds per image generation compared to 27.79 seconds for Regional-Prompting. This demonstrates that our two novel masks—Text-Focus Attention Mask ($M_{\mathrm{focus}}$) and Context-Expansion Attention Mask ($M_{\mathrm{expn}}$)—enable effective and efficient attention control for visual text generation.

\begin{table}[!t]\centering

\footnotesize
\begin{adjustbox}{max width=\columnwidth}
\begin{tabular}{l|rr|rrr|rr}
\toprule
Method &Acc. &NED &Qual. &Aesth. \\\midrule
AnyText &0.096 &0.442 &4.128 &2.990 \\
GlyphControl &0.630 &0.901 &3.935 &2.884 \\
TextDiffuser2 &0.552 &0.860 &3.463 &2.488 \\
EasyText &0.159 &0.484 &4.380 &3.361 \\
DCText (Ours) &0.387 &0.751 &4.737 &3.904 \\
\bottomrule
\end{tabular}

\end{adjustbox}
\caption{\textbf{Quantitative comparison between training-based baselines.} Results are averaged over three single-sentence datasets (ChineseDrawText, DrawTextCreative, TMDBEval500).}
\label{tab:train_single}

\end{table}

\section{Additional Ablation Studies}
\label{sec-b:Additional Ablation Studies}

\subsection{Text-Focus Attention Mask Design}
\label{sec-b1:Text-Focus Attention Mask Design}
\begin{figure*}[!t]
  \centering
    \includegraphics[width=0.9\textwidth]{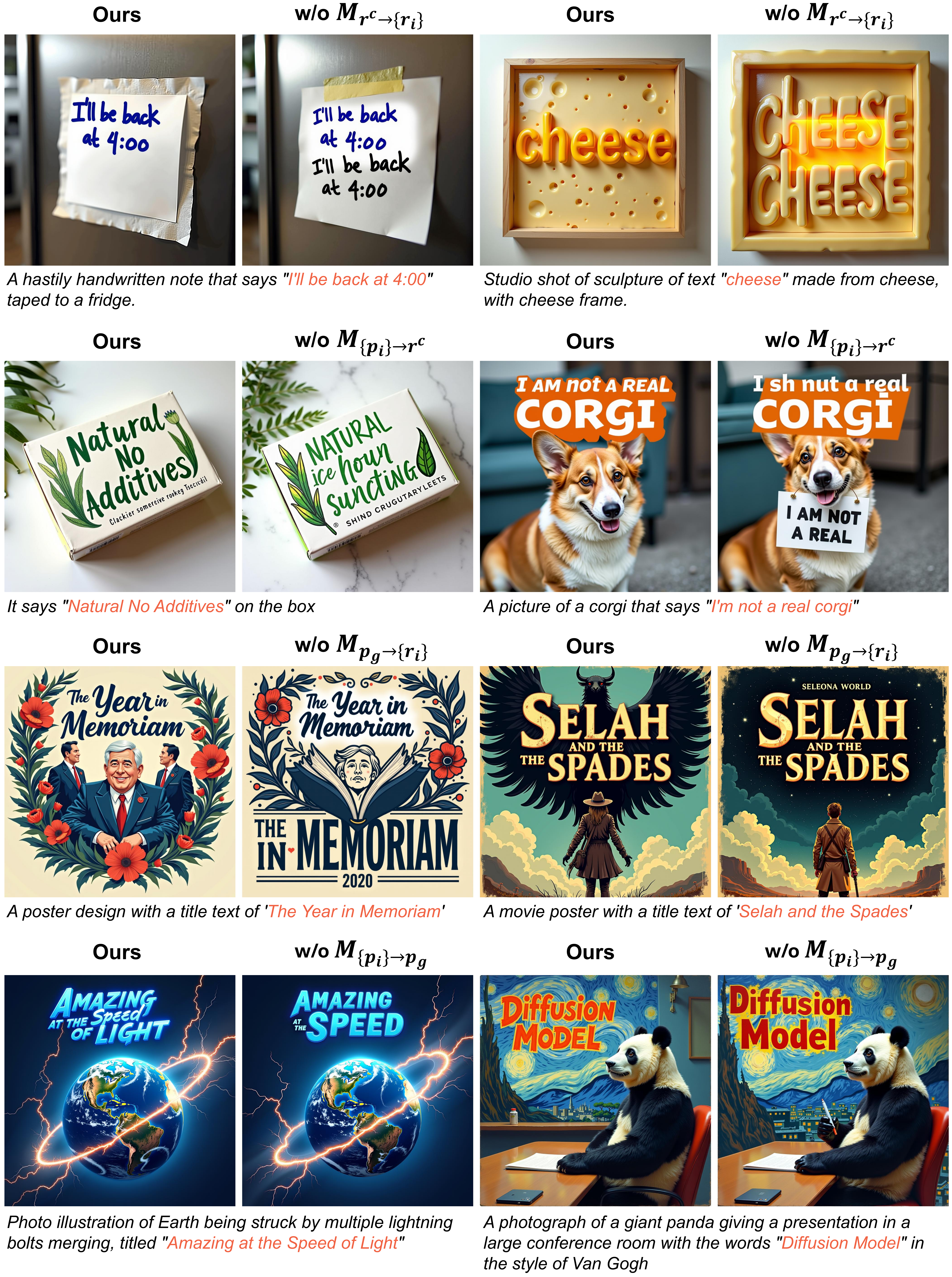} 
    \caption{\textbf{Ablation study for the text-focus attention mask design.} In each pair, the right image shows the result without the corresponding partial mask, and the left image shows the result with it applied.}
    \label{fig:focus_mask_design}
\end{figure*}

To construct $M_{\mathrm{focus}}$, we take the union of four partial masks ($M_{r^c \to \{r_i\}}, M_{\{p_i\} \to r^c}, M_{p_g \to \{r_i\}}, M_{\{p_i\} \to p_g}$), each of which enables a specific directional attention flow. To evaluate the contribution of each component, we conduct an ablation study by selectively excluding individual partial masks. The results are shown in~\cref{fig:focus_mask_design} and~\cref{tab: focus_mask_design}.
When the attention flow from the background region to the textual regions is disabled (\ie{} w/o $M_{r^c \to \{r_i\}}$), duplicated text tends to appear in the background, and the transition between background and target regions becomes unnatural. The awkward bright areas of the first row in~\cref{fig:focus_mask_design} illustrate this issue.
Disabling the attention flow from textual prompts to the background region (w/o $M_{\{p_i\} \to r^c}$) often causes incorrect text to be generated, significantly lowering text accuracy.
When the attention from the global prompt to the textual regions is blocked (w/o $M_{p_g \to \{r_i\}}$), irrelevant text tends to appear. On the other hand, removing the attention from textual prompts to the global prompt (w/o $M_{\{p_i\} \to p_g}$) produces text that is less stylistically aligned with the overall image.
Overall, incorporating all four directional attention flows results in the highest text accuracy and consistently high image quality.

\subsection{Text-Focus Denoising Steps}
\label{sec-b2:Text-Focus Denoising Steps}
\cref{fig:focus_mask_steps} and~\cref{tab: text-focus_mask_steps} present the results of an ablation study on varying the number of denoising steps $T_\mathrm{focus}$ during which the text-focus attention mask $M_\mathrm{focus}$ is applied, while keeping $T_\mathrm{init}$ and $T_\mathrm{expn}$ fixed.
As shown in the figure, when $T_\mathrm{focus} = 0$, that is, when $M_\mathrm{focus}$ is not applied, the model fails to focus on the designated region, often producing region-irrelevant or entirely missing text. As $T_\mathrm{focus}$ increases, alignment between the generated text and the target region improves. However, excessive values of $T_\mathrm{focus}$ prevent regions from attending to the background for extended periods, leading to more noticeable boundaries between regions and their surroundings, and, as shown in the table, even causing a decline in text accuracy.

\subsection{Context-Expansion Denoising Steps}
\label{sec-b3:Context-Expansion Denoising Steps}
\begin{figure*}[!t]
  \centering
    \includegraphics[width=0.95\textwidth]{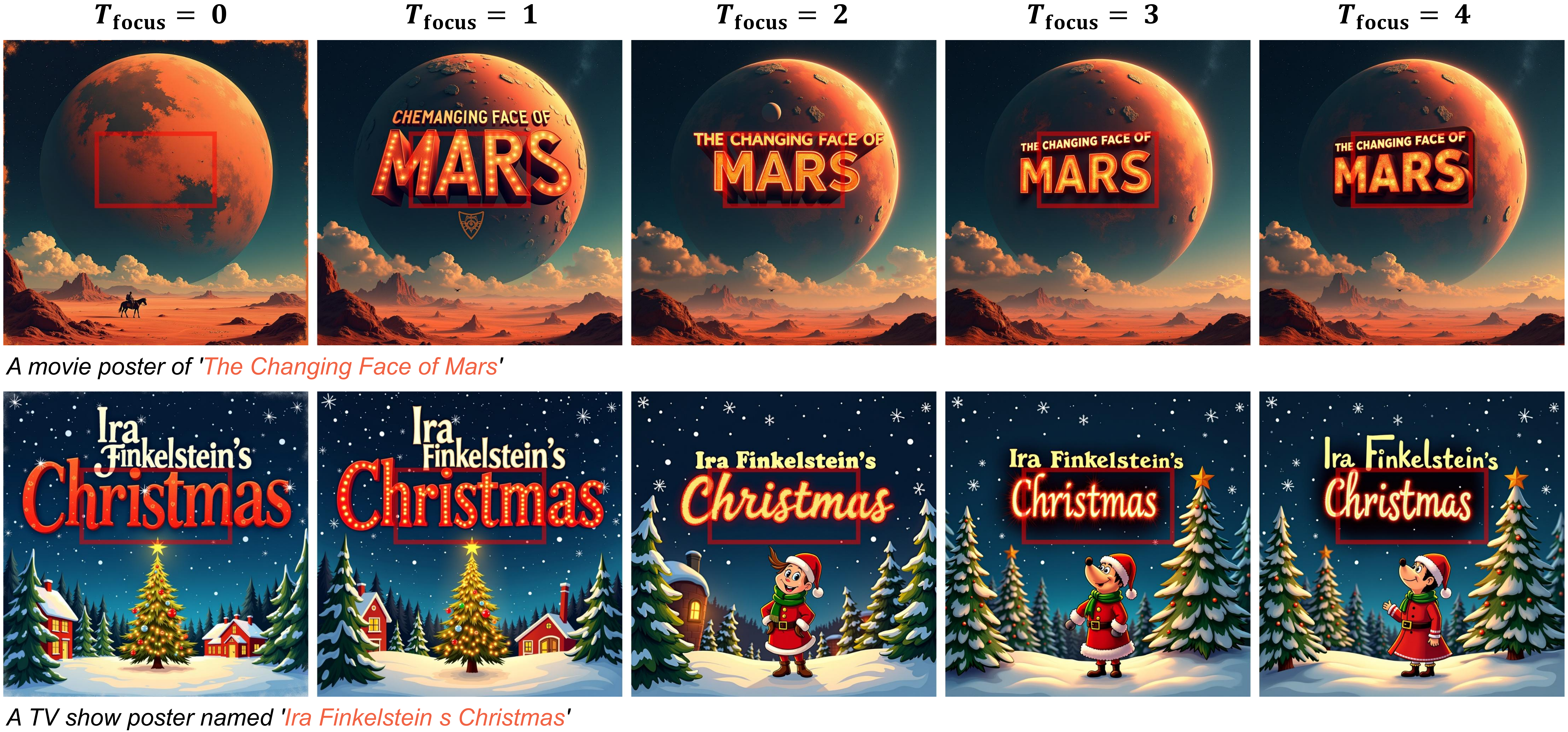} 
    \caption{\textbf{Ablation study for $\bm{T}_{\bm{\mathrm{focus}}}$ steps.} Qualitative results  for varying $T_\mathrm{focus}$, with $T_\mathrm{init} = 1$ and $T_\mathrm{expn} = 2$ fixed.}
    \label{fig:focus_mask_steps}
\end{figure*}

\begin{figure*}[!t]
  \centering
    \includegraphics[width=0.95\textwidth]{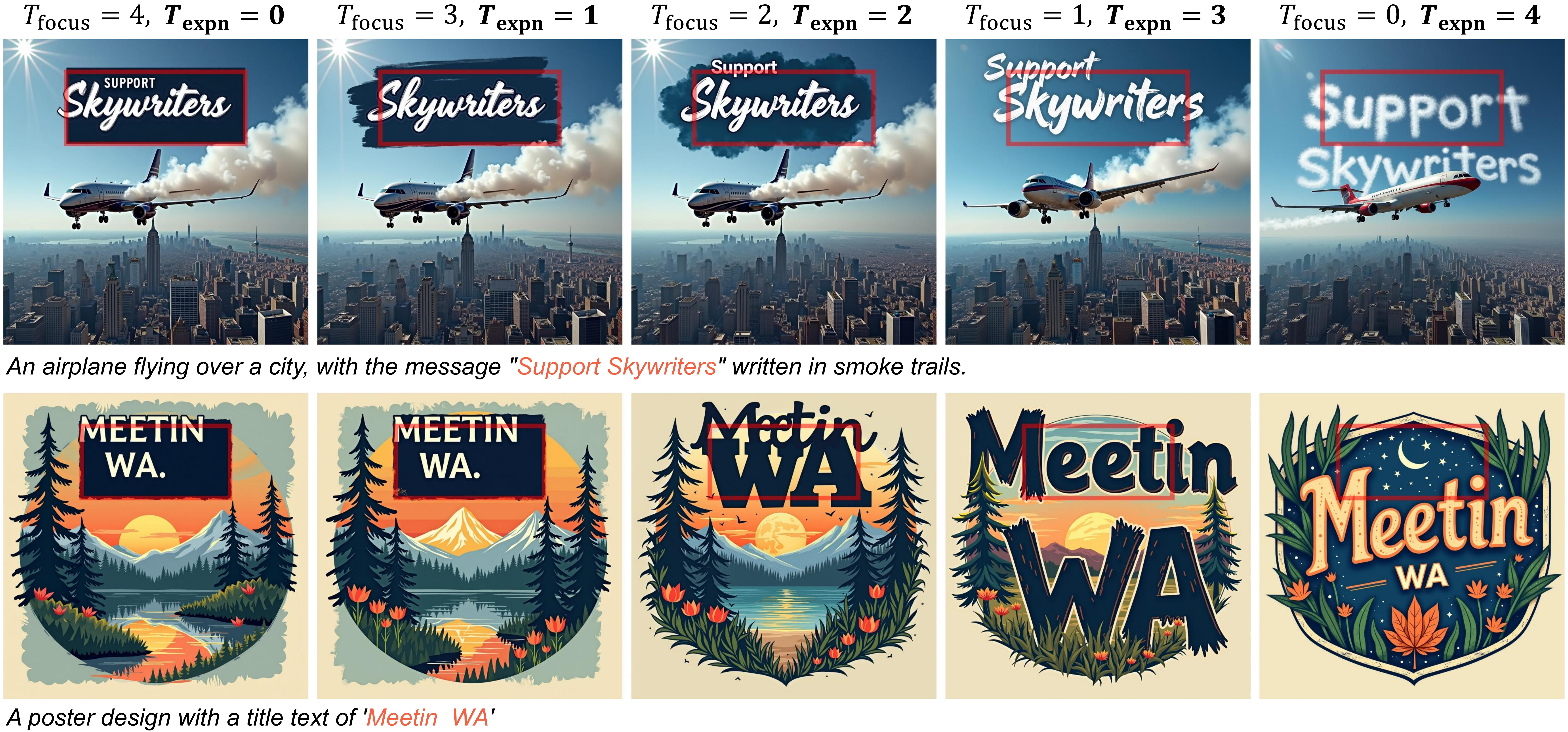} 
    \caption{\textbf{Ablation study for $\bm{T}_{\bm{\mathrm{expn}}}$ steps.} Qualitative results for varying $T_\mathrm{expn}$, where $T_\mathrm{focus}$ is reduced accordingly under a fixed total number of steps, with $T_\mathrm{init} = 1$ fixed.}
    \label{fig:expn_mask_steps}
\end{figure*}

To assess the contribution of the context-expansion attention mask $M_\mathrm{expn}$, we conduct an ablation study varying the number of denoising steps allocated to the text-focus phase ($T_\mathrm{focus}$) and the context-expansion phase ($T_\mathrm{expn}$), keeping the total number of $T_\mathrm{focus} + T_\mathrm{expn}$ steps fixed (\ie{} gradually substituting $M_\mathrm{focus}$ with $M_\mathrm{expn}$). \cref{fig:expn_mask_steps} and~\cref{tab: expn-mask-steps} show qualitative and quantitative results under different allocations of these steps.
When $T_\mathrm{expn} = 0$ (leftmost column), the target text is accurately aligned within the designated region, but the lack of attention to surrounding context results in sharp and unnatural boundaries between the region and background. As more steps are allocated to context-expansion, this boundary effect is gradually alleviated, leading to more visually natural results. However, when $T_\mathrm{expn}$ becomes too dominant, information within the region starts to leak outward, leading to text generation that is no longer confined to the intended region (similar to the failure cases shown in the leftmost examples of~\cref{fig:focus_mask_steps}).
These results highlight the importance of a balanced sequential application of $T_\mathrm{focus}$ and $T_\mathrm{expn}$—where $T_\mathrm{focus}$ helps localize the text within the target region, and $T_\mathrm{expn}$ promotes natural integration into the full image. \cref{tab: expn-mask-steps} further supports this finding: both overly strong text-focus attention (first row) and excessive context-expansion (last row) lead to performance drops, while a balanced allocation yields the balanced high performance across all metrics.

\subsection{Localized Noise Initialization Steps}
\label{sec-b4:Localized Noise Initialization Steps}
\begin{figure*}[!t]
  \centering
    \includegraphics[width=0.6\textwidth]{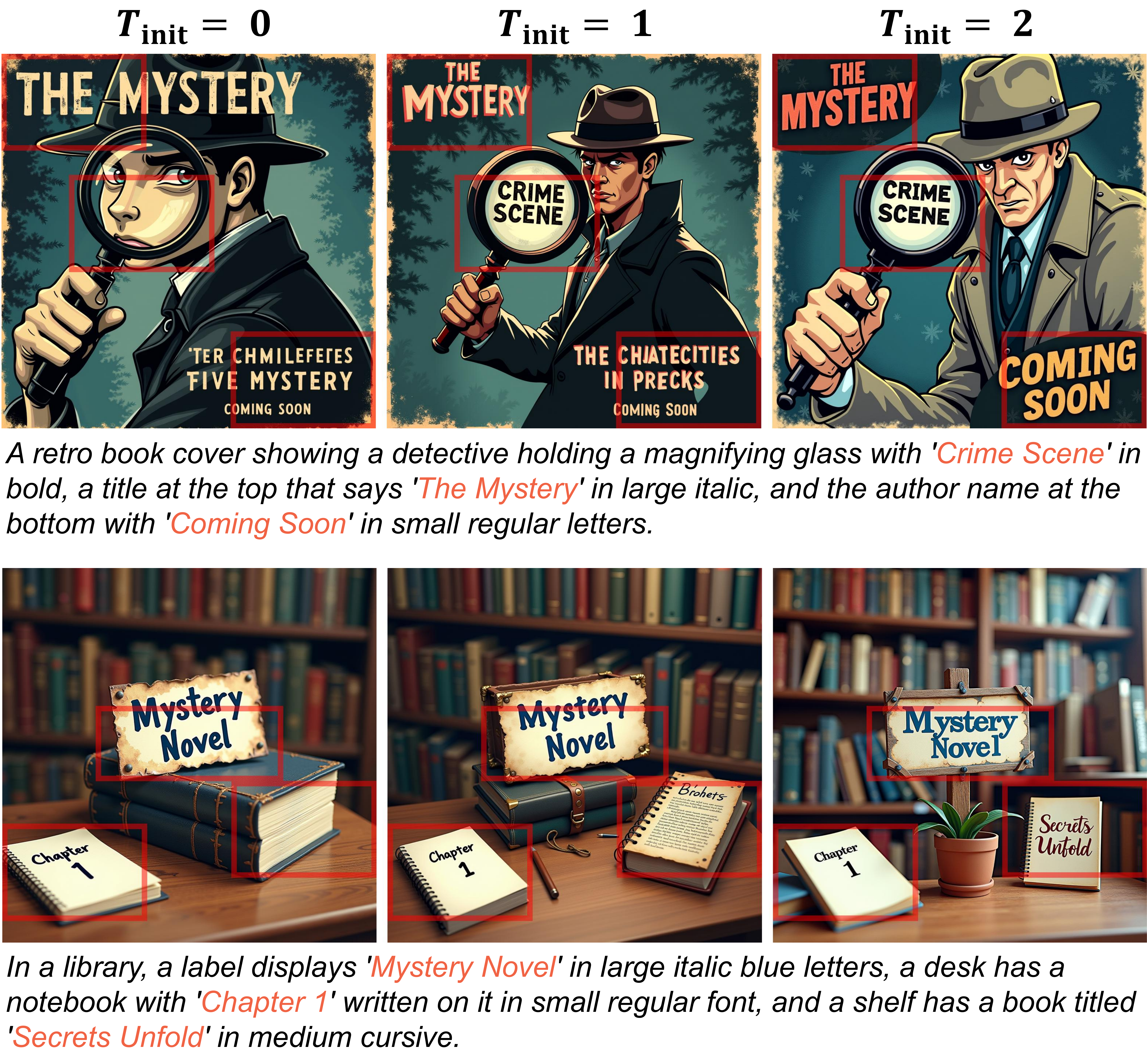} 
    \caption{\textbf{Ablation study for $\bm{T}_{\bm{\mathrm{init}}}$ steps.} Qualitative results for varying $T_\mathrm{init}$, with $T_\mathrm{focus} = 3$ and $T_\mathrm{expn} = 2$ fixed.}
    \label{fig:init_steps}
\end{figure*}

As shown in~\cref{fig:init_steps} and~\cref{tab: init_steps}, increasing $T_\mathrm{init}$ improves text alignment within textual regions and enhances text accuracy. However, since this approach creates an initial latent with uneven noise levels between region and background, we observe that setting $T_\mathrm{init} > 2$ leads to image collapse under our experimental setup with 24 denoising steps.

\begin{table}[!tbp]\centering

\footnotesize
\begin{adjustbox}{max width=\columnwidth}
\begin{tabular}{l|r|rr|rrr|rr}
\toprule
$n$ & Method & Acc. & NED & Qual. & Aesth.  \\
\midrule\midrule
\multirow{4}{*}{$5$}
&AnyText &0.065 &0.275 &4.160 &2.647 \\
&GlyphControl &0.490 &0.722 &3.679 &2.369 \\
&TextDiffuser2 &0.028 &0.241 &3.847 &2.565 \\
&EasyText &0.405 &0.744 &4.049 &2.502 \\
&\textbf{DCText (Ours)} &\textbf{0.693} &\textbf{0.860} &\textbf{4.697} &\textbf{3.569} \\
\midrule
\multirow{4}{*}{$4$}
&AnyText &0.052 &0.269 &4.324 &2.815 \\
&GlyphControl &0.507 &0.729 &3.867 &2.512 \\
&TextDiffuser2 &0.081 &0.321 &3.869 &2.493 \\
&EasyText &0.454 &0.759 &4.235 &2.717 \\
&\textbf{DCText (Ours)} &\textbf{0.760} &\textbf{0.892} &\textbf{4.745} &\textbf{3.659} \\
\midrule
\multirow{4}{*}{$3$}
&AnyText &0.054 &0.261 &4.281 &2.781 \\
&GlyphControl &0.610 &0.795 &3.853 &2.521 \\
&TextDiffuser2 &0.252 &0.508 &3.785 &2.439 \\
&EasyText &0.433 &0.762 &4.266 &2.747 \\
&\textbf{DCText (Ours)} &\textbf{0.768} &\textbf{0.906} &\textbf{4.735} &\textbf{3.709} \\
\midrule
\multirow{4}{*}{$2$}
&AnyText &0.052 &0.275 &4.386 &2.857 \\
&GlyphControl &0.692 &0.862 &4.009 &2.624 \\
&TextDiffuser2 &0.528 &0.729 &3.755 &2.438 \\
&EasyText &0.460 &0.791 &4.363 &2.870 \\
&\textbf{DCText (Ours)} &\textbf{0.792} &\textbf{0.923} &\textbf{4.791} &\textbf{3.726} \\
\bottomrule
\end{tabular}
\end{adjustbox}

\caption{\textbf{Quantitative comparison between training-based baselines.} Comparison results on the CVTG-Style dataset across different numbers of sentences ($n$).}
\label{tab:train_multi}

\end{table}

\begin{table}[!t]\centering
\resizebox{\columnwidth}{!}{ 
\begin{tabular}{l|rr|rrrr}\toprule
Mask &Acc. &NED &CLIP &Qual. &Aesth. \\\midrule
w/o $M_{r^c \to \{r_i\}}$ &0.275 &0.683 &0.347 &4.721 &3.878 \\
w/o $M_{\{p_i\} \to r^c}$ &0.266 &0.666 &0.345 &4.735 &3.885 \\
w/o $M_{p_g \to \{r_i\}}$ &0.330 &0.721 &\textbf{0.350} &4.733 &3.890 \\
w/o $M_{\{p_i\} \to p_g}$ &0.347 &0.732 &0.349 &\textbf{4.746} &\textbf{3.921} \\
$\bm{M}_{\bm{\mathrm{focus}}}$ &\textbf{0.387} &\textbf{0.751} &0.349 &4.737 &3.904 \\
\bottomrule
\end{tabular}
}

\caption{\textbf{Ablation study for $\bm{M}_{\bm{\mathrm{focus}}}$ design.} Each row reports the result when one of the partial masks (defined in~\cref{Attention Mask Control}) is removed, evaluated on the single-sentence datasets.}
\label{tab: focus_mask_design}
\end{table}
\begin{table}[!t]\centering

\begin{adjustbox}{max width=0.8\columnwidth}
\begin{tabular}{c|rr|rrrr}\toprule
$\bm{T}_{\bm{\mathrm{focus}}}$ &Acc. &NED &CLIP &Qual. &Aesth. \\\midrule
0 &0.273 &0.626 &0.344 &4.739 &3.888 \\
1 &0.316 &0.701 &0.348 &\textbf{4.740} &\textbf{3.905} \\
\textbf{2} &\textbf{0.387} &\ul{0.751} &\ul{0.349} &\ul{4.737} &\ul{3.904} \\
3 &\ul{0.372} &\textbf{0.757} &\textbf{0.350} &4.728 &3.894 \\
4 &0.337 &0.746 &0.349 &4.717 &3.875 \\
\bottomrule
\end{tabular}
\end{adjustbox}

\caption{\textbf{Ablation study for $\bm{T}_{\bm{\mathrm{focus}}}$ steps.} Quantitative results for different values of $T_\mathrm{focus}$, evaluated on the single-sentence datasets.}
\label{tab: text-focus_mask_steps}
\end{table}

\section{Broader Applications of DCText}
\label{sec-c:Broader Applications of DCText}

\subsection{General Object}
\label{sec-c1:General Object}
While \method{} is primarily designed to address the challenging task of rendering long or multiple texts, its core strategy generalizes effectively to broader visual generation tasks. To evaluate this generality, we apply \method{} to the GenEval~\cite{ghosh2023geneval} benchmark, which focuses on the compositional generation of general objects.
In this experiment, we follow the original \method{} pipeline as-is, but modify the GPT-4o~\cite{hurst2024gpt} instructions for constructing textual prompts and regions. For textual prompts, we extract the target object from the global prompt and generate an object-centric prompt that includes a description aligned with the global context. For textual regions, we revise the original text-based instructions into object-based ones.
We use the same denoising schedule as in the text generation setup: $(T_{\mathrm{init}},T_{\mathrm{focus}},T_{\mathrm{expn}})=(1,2,2)$ for single-object generation and $(2,3,2)$ for multi-object generation.

\begin{figure*}[!t]
  \centering
    \includegraphics[width=0.95\textwidth]{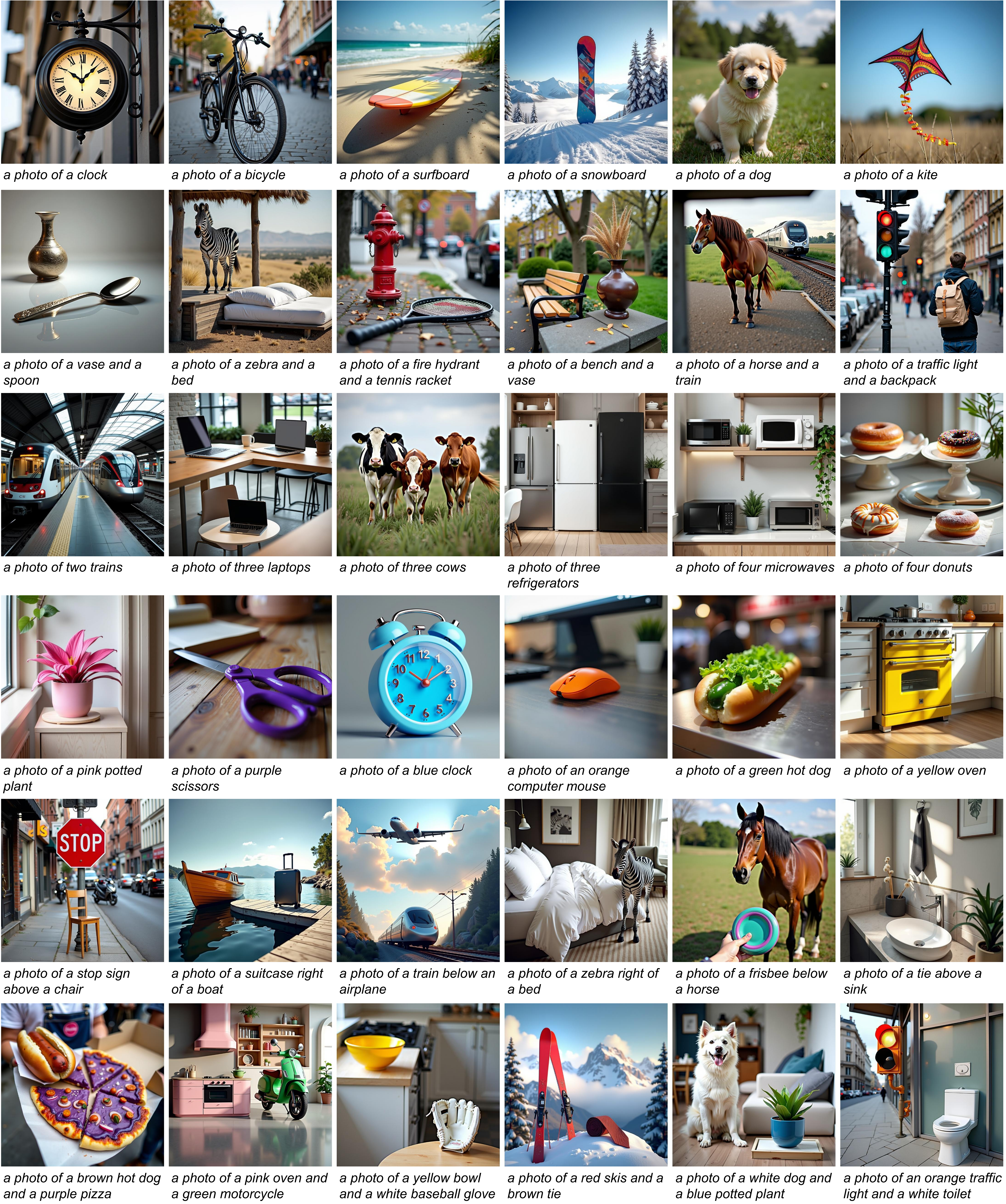} 
    \caption{\textbf{Qualitative samples on the GenEval benchmark.} DCText-generated samples on the GenEval benchmark. Rows correspond to Single Object, Two Object, Counting, Colors, Position, and Attribution Binding tasks.}
    \label{fig:geneval}
\end{figure*}

\cref{fig:geneval} shows qualitative results, and~\cref{tab:geneval} summarizes quantitative comparisons.
For evaluation, we generate four samples per prompt across all 553 prompts in the benchmark. As shown, \method{} significantly outperforms the base model Flux, improving the overall GenEval score from 0.66 to 0.78. These results demonstrate the strong generalization capability of \method{} beyond text rendering.

\subsection{Stable Diffusion 3.5}
\label{sec-c2:Stable Diffusion 3.5}
We further evaluate the performance of \method{} on another Multi-Modal Diffusion Transformer model, Stable Diffusion 3.5 Large (SD3.5-L)~\cite{sd3}. Following the same experimental setup as in the main paper, we also compare \method{} against the same three baselines: SD3.5-L (the base model), AMO Sampler~\cite{hu2025amosamplerenhancingtext}, and TextCrafter~\cite{du2025textcrafter}. For fair comparison, we use a fixed number of 28 denoising steps across all methods, while keeping all other configurations at their respective defaults. As the AMO Sampler does not provide an official implementation for SD3.5, we re-implement it ourselves.

\begin{figure*}[!t]
  \centering
    \includegraphics[width=0.923\textwidth]{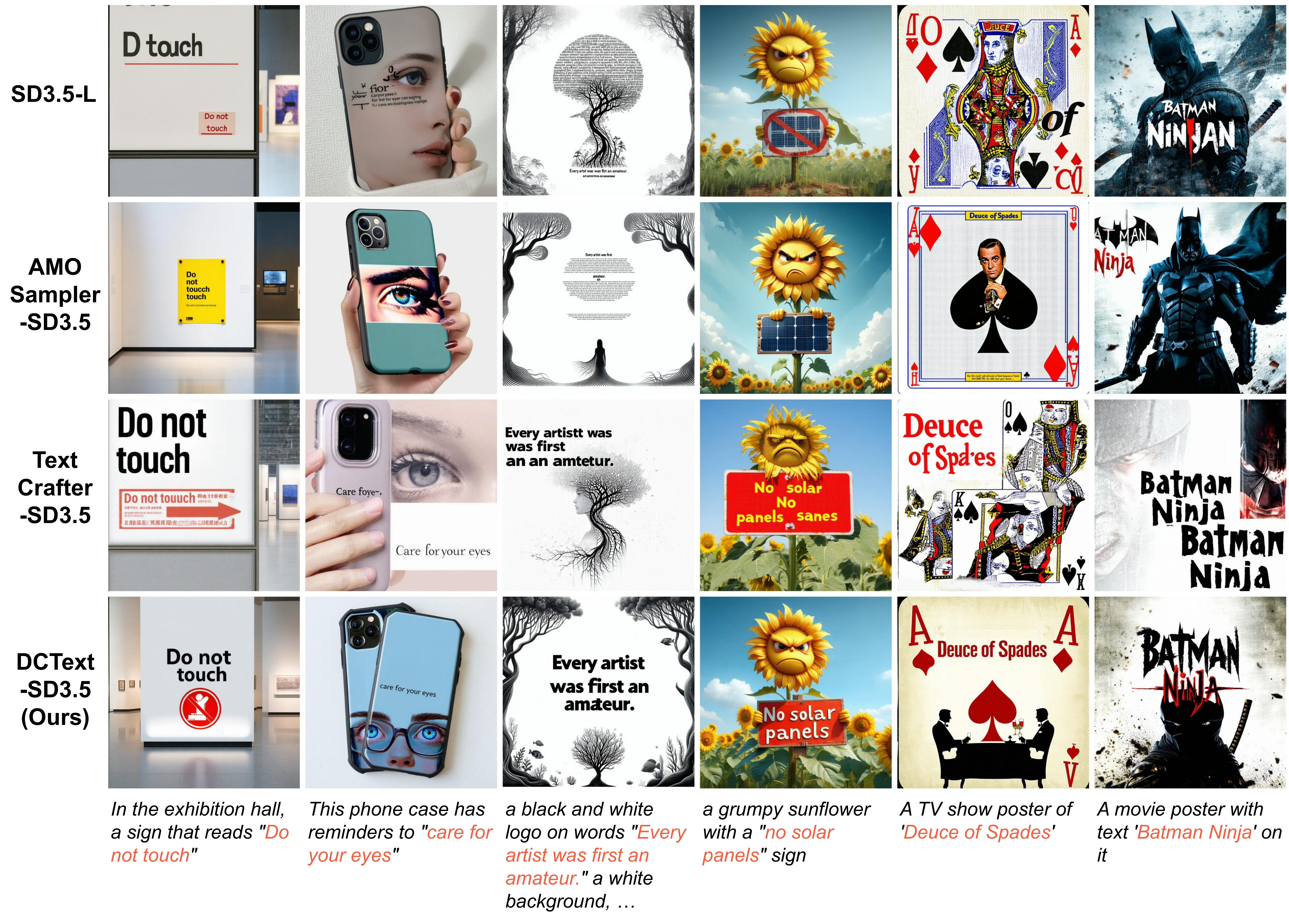}
    \caption{\textbf{Qualitative comparison between SD3.5-based baselines.} Samples generated by each method using the SD3.5-L.}
    \label{fig:single_sd35}
\end{figure*}

As shown in~\cref{fig:single_sd35}, \method{} consistently produces accurate and coherent text aligned with the overall image context. In contrast, SD3.5-L and AMO Sampler often fail to render any text or generate inaccurate content, while TextCrafter tends to produce duplicated text and unnatural region boundaries.
\cref{tab:single_comparison_sd35} presents the quantitative results on three single-sentence datasets, where all baselines generate three samples per prompt using the same random seed. Consistent with the Flux-based results in the main paper, \method{} achieves the best performance across most metrics, including text accuracy and image quality.





\begin{table}[!t]\centering

\begin{adjustbox}{max width=0.95\columnwidth}
\begin{tabular}{c|rr|rrrr}\toprule
$\bm{T}_{\bm{\mathrm{expn}}} (T_{\mathrm{focus}})$ &Acc. &NED &CLIP &Qual. &Aesth. \\\midrule
0 (4) &0.289 &0.681 &0.347 &4.716 &3.829 \\
1 (3) &0.339 &0.723 &\ul{0.349} &\textbf{4.739} &3.895 \\
\textbf{2 (2)} &\textbf{0.387} &\textbf{0.751} &\ul{0.349} &\ul{4.737} &\ul{3.904} \\
3 (1) &\ul{0.340} &\ul{0.750} &\textbf{0.350} &4.731 &\textbf{3.905} \\
4 (0) &0.336 &0.746 &\ul{0.349} &4.725 &3.889 \\
\bottomrule
\end{tabular}
\end{adjustbox}

\caption{\textbf{Ablation study for $\bm{T}_{\bm{\mathrm{expn}}}$ steps.} Quantitative results under different allocations of $T_\mathrm{expn}$ and $T_\mathrm{focus}$ (values in parentheses). As $T_\mathrm{expn}$ increases, $T_\mathrm{focus}$ is reduced accordingly, evaluated on the single-sentence datasets.}
\label{tab: expn-mask-steps}
\end{table}

\begin{table}[!t]\centering

\begin{adjustbox}{max width=0.8\columnwidth}
\begin{tabular}{c|rr|rrrr}\toprule
$\bm{T}_{\bm{\mathrm{init}}}$ &Acc. &NED &CLIP &Qual. &Aesth. \\\midrule
0 &0.600 &0.798 &0.347 &4.640 &3.538 \\
1 &0.717 &0.878 &0.348 &4.702 &3.653 \\
\textbf{2} &\textbf{0.753} &\textbf{0.895} &\textbf{0.349} &\textbf{4.742} &\textbf{3.666} \\
\bottomrule
\end{tabular}
\end{adjustbox}

\caption{\textbf{Ablation study for $\bm{T}_{\bm{\mathrm{init}}}$ steps.} Quantitative results for different values of $T_\mathrm{init}$, evaluated on the multi-sentence dataset.}
\label{tab: init_steps}
\end{table}

\definecolor{color_blue}{RGB}{102, 181, 255}
\definecolor{color_green}{RGB}{51, 153, 51}

\begin{table*}[!t]
    \centering
    
    \resizebox{\linewidth}{!}{
    \begin{tabular}{l|c|cccccc}
    \toprule
    \textbf{Model} & \textbf{Overall} & \textbf{Single Obj.} & \textbf{Two Obj.} & \textbf{Counting} & \textbf{Colors} & \textbf{Position} & \textbf{Attr. Binding} \\
    \midrule
    \multicolumn{8}{c}{\textit{Diffusion Models}} \\
    \midrule
    LDM~\cite{rombach2022high}  & 0.37 & 0.92 & 0.29 & 0.23 & 0.70 & 0.02 & 0.05 \\
    SD1.5~\cite{rombach2022high}  & 0.43 & 0.97 & 0.38 & 0.35 & 0.76 & 0.04 & 0.06 \\
    SD2.1~\cite{rombach2022high}  & 0.50 & 0.98 & 0.51 & 0.44 & \colorbox{color_blue!20}{0.85} & 0.07 & 0.17 \\
    SD-XL~\cite{podell2023sdxl}  & 0.55 & 0.98 & 0.74 & 0.39 & \colorbox{color_blue!20}{0.85} & 0.15 & 0.23 \\
    DALLE-2~\cite{ramesh2022hierarchical}  & 0.52 & 0.94 & 0.66 & 0.49 & 0.77 & 0.10 & 0.19 \\
    DALLE-3~\cite{betker2023improving}  & 0.67 & 0.96 & 0.87 & 0.47 & 0.83 & 0.43 & 0.45 \\
    \midrule
    \multicolumn{8}{c}{\textit{Flow Matching Models}} \\
    \midrule
    FLUX.1 Dev~\cite{flux1-dev}  & 0.66 & 0.98 & 0.81 & \colorbox{color_green!20}{0.74} & 0.79 & 0.22 & 0.45 \\
    SD3.5-M~\cite{sd3} & 0.63 & 0.98 & 0.78 & 0.50 & 0.81 & 0.24 & 0.52 \\
    SD3.5-L~\cite{sd3}  & 0.71 & 0.98 & 0.89 & 0.73 & 0.83 & 0.34 & 0.47 \\
    SANA-1.5 4.8B~\cite{xie2025sana} & \colorbox{color_blue!20}{0.81} & \colorbox{color_green!20}{0.99} & \colorbox{color_blue!20}{0.93} & \colorbox{color_blue!20}{0.86} & \colorbox{color_green!20}{0.84} & \colorbox{color_green!20}{0.59} & \colorbox{color_blue!20}{0.65} \\
    \midrule
    \textbf{DCText (Ours)} & \colorbox{color_green!20}{0.78} & \colorbox{color_blue!20}{1.00} & \colorbox{color_green!20}{0.90} & {0.51} & \colorbox{color_green!20}{0.84} & \colorbox{color_blue!20}{0.82} & \colorbox{color_green!20}{0.61}  \\
    \bottomrule
    \end{tabular}
    }

    \caption{\textbf{Quantitative comparison on the GenEval benchmark.} We highlight the best scores in \colorbox{color_blue!20}{blue} and second-best in \colorbox{color_green!20}{green}. Results for all baseline models are adopted from Flow-GRPO \cite{liu2025flow}. Obj.: Object; Attr.: Attribution.}
    \label{tab:geneval}
\end{table*}


\section{Limitation}
\label{sec-d:Limitation}
Our method relies on Flux’s reliable short-text generation capability.
If the textual prompt fails to generate the target text from the noise corresponding to the textual region, our method may not render the text correctly in that area.
In addition, for our method to operate effectively, glyph-level features are expected to emerge before the Text-Focus denoising phase.
This is because, after Text-Focus denoising, attention expands to the background region, followed by global denoising with full attention. While Flux typically forms coarse glyph structures during the early denoising steps, it occasionally fails to produce recognizable glyph features during this phase.

\begin{figure*}[!t]
  \centering
    \includegraphics[width=0.95\textwidth]{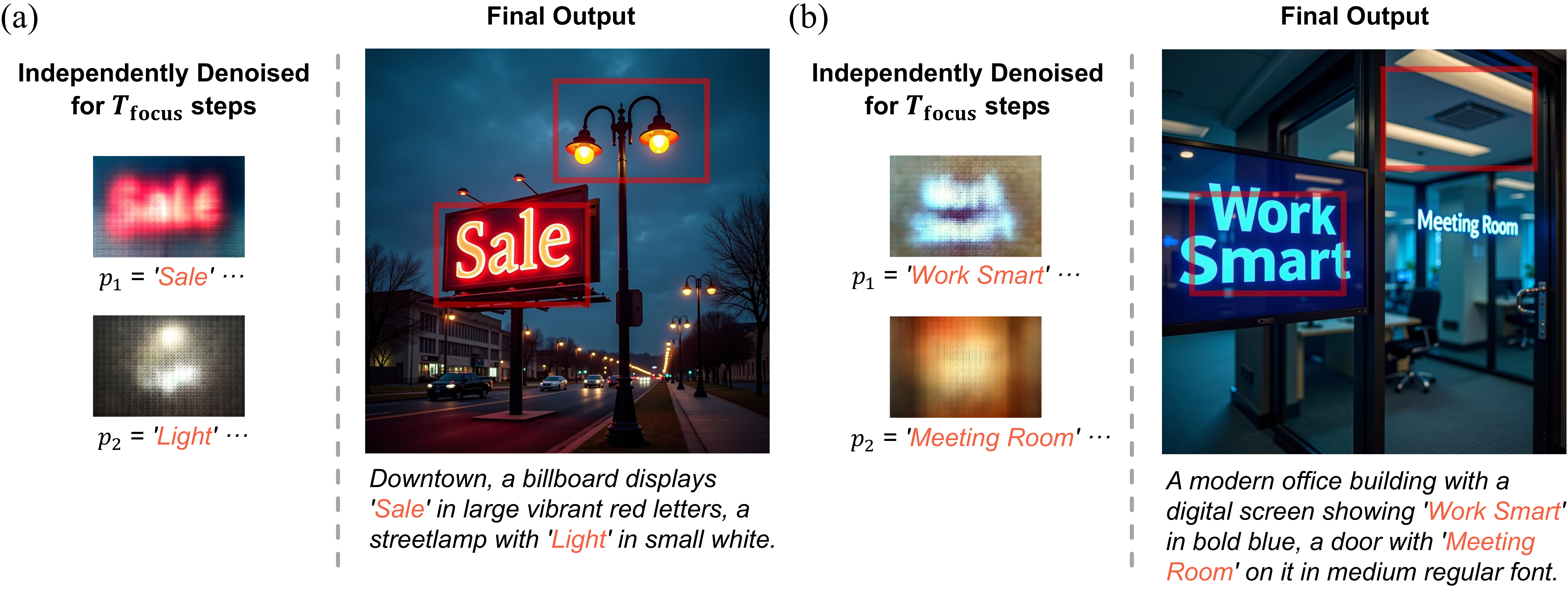}
    \caption{\textbf{Failure Cases.} Each region is extracted from the initial noise used to generate the final image (right) and denoised for $T_{\mathrm{focus}}$ steps using the corresponding textual prompts (left). (a) The prompt $p_1$ leads to clear glyph-like features, but not $p_2$. As a result, only \textit{Sale} appears in the final image. (b) Similar case where the region for $p_2$ fails to form glyphs early on. Nevertheless, the global prompt allows \textit{Meeting Room} to appear during global denoising.}
    \label{fig:limitation}
\end{figure*}

\cref{fig:limitation}a illustrates such a case. The left images show intermediate results obtained by independently denoising each region for $T_{\mathrm{focus}}$ steps (with $T_{\mathrm{init}}$ set to 0 for simplicity). 
In the image for $p_1$, features resembling the word \textit{sale} begin to emerge, whereas in the image for $p_2$, the model generates features related to the object light rather than the text \textit{light}. In such cases, our method often fails to render the target text in the corresponding region, as shown in the final output on the right.

However, such failures are often compensated for during global denoising.
As in~\cref{fig:limitation}a,~\cref{fig:limitation}b also shows that no glyph-like features appear in the image for $p_2$, leading to missing text in that region of the final image.
Nevertheless, since the global prompt includes the phrase corresponding to $p_2$, the final image still successfully generates the text \textit{Meeting Room}.

\begin{table}[t]\centering

\footnotesize
\begin{adjustbox}{max width=\columnwidth}
\begin{tabular}{l|rr|rrrrr}
\toprule
Method & Acc. & NED & CLIP & Qual. & Aesth. \\
\midrule
SD3.5-L & 0.264 & 0.654 & 0.362 & 4.448 & 3.643 \\
AMO-SD3.5 & 0.351 & 0.685 & 0.360 & 4.496 & 3.715 \\
TextCrafter-SD3.5 & 0.241 & 0.707 & \textbf{0.366} & 4.326 & 3.507 \\
\textbf{DCText-SD3.5 (ours)} & \textbf{0.359} & \textbf{0.742} & 0.360 & \textbf{4.618} & \textbf{3.728}  \\
\bottomrule
\end{tabular}

\end{adjustbox}
\caption{\textbf{Quantitative comparison between SD3.5-based baselines.} Results are averaged over three single-sentence datasets. Since the official implementation of AMO-SD3.5 is not available, we implemented it ourselves.}
\label{tab:single_comparison_sd35}

\end{table}

\section{Experimental Details}
\label{sec-e:Experimental Details}

\subsection{Implementation}
\label{sec-e1:Implementation}
For the ChineseDrawText~\cite{ma2023glyphdraw}, DrawTextCreative~\cite{liu2022character}, and TMDBEval500~\cite{chen2023textdiffuser} datasets, we generate both textual prompts and textual regions using GPT‑4o~\cite{hurst2024gpt}. Textual prompts are constructed following the instruction in~\cref{tab:textual_prompt}. For each sentence contained in the prompt, we produce a description and format the result as: \textit{`Rendering word: ``\{sentence\}"\textbackslash n Description: \{description\}'}.
Textual regions are constructed according to the bounding box generation instructions outlined in~\cref{tab:textual_region}.
In the Localized Noise Initialization process, we set the weighting factor $\alpha = 0.7$. During denoising, we use a guidance scale of $5.0$. Ours attention masks are applied to all MM‑DiT blocks, including both double- and single-stream variants. For pooled textual embeddings, we average the embeddings obtained from all textual prompts, including the global prompt.
The text accuracy for multiple sentences is evaluated using GPT-based recognition, following the instructions provided in~\cref{tab:gpt_eval}.

\subsection{Human Evaluation}
\label{sec-e2:Human Evaluation}
\begin{figure*}[!t]
  \centering
    \includegraphics[width=0.95\textwidth]{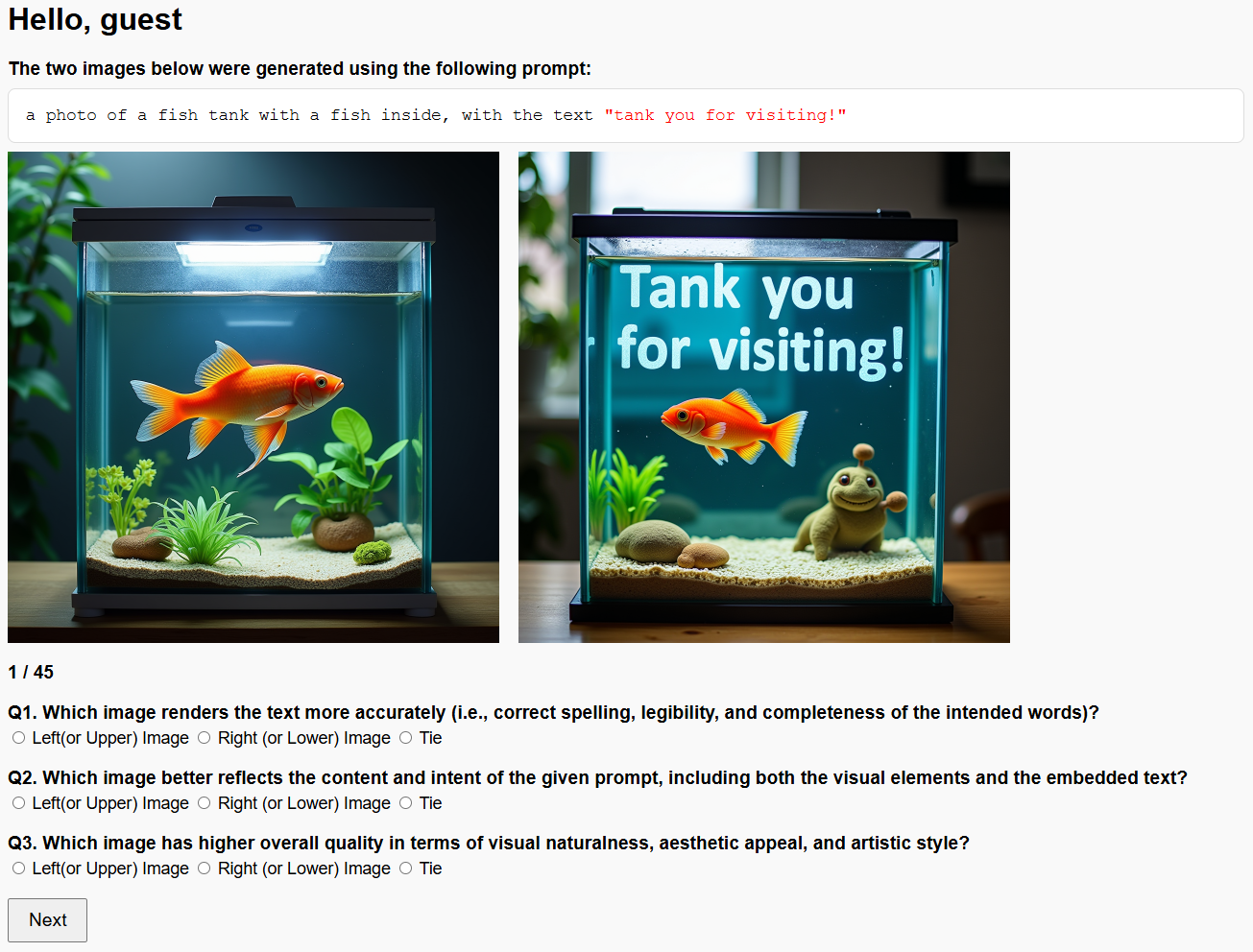} 
    \caption{\textbf{Human evaluation interface.} For each prompt, evaluators perform a pairwise comparison of two generated images, assessing them on text accuracy, prompt alignment, and image quality.}
    \label{fig:humanCapture}
\end{figure*}

We conduct our human evaluation using a pairwise comparison (A/B test) protocol. In each test, participants are shown two images: one from our proposed model, \method{}, and one from a randomly selected baseline (Flux~\cite{flux1-dev}, AMO Sampler~\cite{hu2025amosamplerenhancingtext}, or TextCrafter~\cite{du2025textcrafter}). To mitigate bias, the display order of the images is randomized. The participants are then asked to choose the superior image based on the following three criteria:

\begin{itemize}
    \item \textbf{Text Accuracy:} \textit{Which image renders the text more accurately (i.e., correct spelling, legibility, and completeness of the intended words)?}
    
    \item \textbf{Prompt Alignment:} \textit{Which image better reflects the content and intent of the given prompt, including both the visual elements and the embedded text?}

    \item \textbf{Image Quality:} \textit{Which image has higher overall quality in terms of visual naturalness, aesthetic appeal, and artistic style?}
\end{itemize}
The evaluation interface is illustrated in~\cref{fig:humanCapture}. 

To assess the significance of user preferences, we perform one-sided binomial tests for each pairwise comparison, excluding ties. \method{} shows statistically significant improvements in text accuracy over all baselines ($p < 0.0001$), and in prompt alignment over AMO Sampler and TextCrafter ($p < 0.001$). For overall image quality, the improvement over TextCrafter is also significant ($p = 0.002$), while those over AMO and FLUX do not reach significance.

\subsection{Abbreviated Prompts}
\label{sec-e3:Abbreviated Prompts}
Due to space constraints in Fig~\ref{fig:thumbnail} and~\ref{fig:multi} of the main paper, we present only abbreviated examples of the global prompts. The complete set of prompts is provided in~\cref{tab:full_prompt}, where the target rendering text is highlighted.


\begin{table*}[!t]
\centering
\begin{tabularx}{\textwidth}{X}
\toprule
You are given a text-to-image generation prompt that includes quoted text. \\[4pt]
Your task is to extract each quoted sentence and generate a visual style description for the text inside the quotation marks.
\begin{itemize}[topsep=4pt]
    \item Describe how the text visually appears in the image, including font style, color, texture, effects, etc.
    \item For each extracted sentence, write a concise and context-aware visual style description.
    \item Do not describe the sentence's position or relative order.
    \item Do not mention any rendering words. Avoid using quotation marks or referring to specific text.
\end{itemize} \\[6pt]

Example: \\[4pt]
\texttt{[} \\[-2pt]
\texttt{\hspace*{2ex}\{} \\[-2pt]
\texttt{\hspace*{4ex}"sentence": "diamonds",} \\[-2pt]
\texttt{\hspace*{4ex}"description": "A sleek, modern sans-serif font in metallic silver."} \\[-2pt]
\texttt{\hspace*{2ex}\}} \\[-2pt]
\texttt{]} \\
\bottomrule
\end{tabularx}
\caption{GPT‑4o instruction for generating sentence descriptions within textual prompts.}
\label{tab:textual_prompt}
\end{table*}



\begin{table*}[!t]
\centering
\begin{tabularx}{\textwidth}{X}
\toprule

You are given a text-to-image prompt with quoted text. \\
Your task is to extract the quoted text and generate a bounding box. \\[6pt]

\textbf{Step-by-step Instructions} \\[6pt]

\textbf{1. Quoted Text Isolation}
\begin{itemize}[topsep=4pt]
    \item Extract the text inside quotation marks only.
    \item Example: \newline Prompt: A sign that says \texttt{"Do not reserve a seat"} $\to$ Use: \texttt{`Do not reserve a seat'}
\end{itemize} \\

\textbf{2. Bounding Box Layout Rules}
\begin{itemize}[topsep=4pt]
    \item The bounding box must be placed in regions where the text is likely to appear, as implied by the prompt.
    \item Bounding boxes must not overlap.
\end{itemize} \\[4pt]

\textbf{3. Bounding Box Calculation}
\begin{itemize}[topsep=4pt]
    \item Output each bounding box as normalized coordinates, meaning all values (x1, y1, x2, y2) are between 0 and 1, representing a fraction of the image width and height.
    \item Consider the number of characters, including spaces and punctuation, for the size of the box.
    \item The height of every bounding box must be \texttt{\{height\}}.
    \item The width of every bounding box must be at least \texttt{\{min\_width\}}.
    \item Final format: \texttt{[x1, y1, x2, y2]}.
\end{itemize} \\

\bottomrule
\end{tabularx}
\caption{GPT‑4o instruction for generating bounding boxes of textual regions.}
\label{tab:textual_region}
\end{table*}

\begin{table*}[!t]
\centering
\begin{tabularx}{\textwidth}{X}
\toprule
Recognize all textual elements in the image as they would be perceived by a human and organize them into accurate, sentence-level units.
\begin{itemize}[topsep=4pt]
\item Split the text based on meaningful sentence boundaries.
\item Each sentence must come from a single region in the image.
\item Do not correct or modify awkward words or phrases.
\item Include a score indicating the visual recognition confidence of each sentence.
\end{itemize} \\[6pt]

Example Output Format (JSON): \\[4pt]

\texttt{[} \\
\texttt{\hspace*{4ex}\{"sentence": "New Specials Every Week", "score": 0.96\},} \\
\texttt{\hspace*{4ex}\{"sentence": "We are OPEN EVERY DAY", "score": 0.91\}} \\
\texttt{]} \\
\bottomrule
\end{tabularx}
\caption{GPT-4o instruction for text recognition in accuracy evaluation.}
\label{tab:gpt_eval}
\end{table*}

\definecolor{color_red}{RGB}{241,99,68}

\begin{table*}[!t]
    \centering
    
    \begin{small}
    \renewcommand{\arraystretch}{1.8}
    \begin{tabularx}{\textwidth}{@{} l X @{}}
    \toprule
    \textbf{Figure} & \textbf{Prompt} \\
    \midrule

    \multirow{12}{*}{\textbf{Figure 1}} 
    & $\bullet$\texttt{ A sprawling financial district at dusk, where the text "\textcolor{color_red}{DCText: Scheduled Attention Masking for Visual Text Generation via Divide-and-Conquer Strategy}" is projected across the mirrored glass surface of a skyscraper. The characters are bold, futuristic sans-serif with a subtle neon blue glow, appearing as if etched into the building façade, reflecting surrounding city lights faintly.} \\
    & $\bullet$\texttt{ A quaint street corner during dusk, with a classic 1950s-style diner sign. The text "\textcolor{color_red}{DCText: Scheduled Attention Masking for Visual Text Generation via Divide-and-Conquer Strategy}" is displayed in a retro script font with glowing red and cream-colored bulbs along the letters, evoking a warm, nostalgic roadside ambiance.} \\
    & $\bullet$\texttt{ A vintage-style parchment sheet with burned edges, where the text "\textcolor{color_red}{DCText: Scheduled Attention Masking for Visual Text Generation via Divide-and-Conquer Strategy}" is hand-painted in an ornate calligraphic style using dark ink with faint gold leaf accents. The imperfect, organic strokes give the title an ancient manuscript appearance.} \\
    & $\bullet$\texttt{ A futuristic night sky above a modern metropolis, where hundreds of synchronized drones form the text "\textcolor{color_red}{DCText: Scheduled Attention Masking for Visual Text Generation via Divide-and-Conquer Strategy}" in the air. Each character is composed of tiny glowing blue lights, forming a perfectly sharp sans-serif display that glimmers softly against the starry sky, while the city below remains dim and distant.} \\
    
    \midrule
    
    \multirow{6}{*}{\textbf{Figure 4}} 
    & $\bullet$\texttt{ On a sunny beach scene, a lifeguard tower sign proclaiming '\textcolor{color_red}{Swimming Area Open}' in large red letters, a beach umbrella with '\textcolor{color_red}{Relax and Enjoy the Sun}' in colorful cursive, a kiosk displaying '\textcolor{color_red}{Beach Gear Rentals}' in medium blue, and a signpost pointing to '\textcolor{color_red}{Surf Lessons Starting at 10 AM}' in bold orange.} \\
    & $\bullet$\texttt{ Charming restaurant exterior showcasing a sign with '\textcolor{color_red}{Family Meals Available}' in large red letters, a window display reading '\textcolor{color_red}{Daily Fresh Catch}' in bold blue letters, a door plaque labeled '\textcolor{color_red}{Welcome Diners}' in green cursive, a patio banner saying '\textcolor{color_red}{Outdoor Seating Open}' in italic large letters, and a menu board displaying '\textcolor{color_red}{Chef Specials Tonight}' in medium bold letters.} \\
    \bottomrule
    
    \end{tabularx}
    \end{small}

    \caption{The full text-to-image prompts for Figure 1 and 4, ordered left to right (Figure 1) and top to bottom (Figure 4).}
    \label{tab:full_prompt}
\end{table*}


\end{document}